\documentclass[conf]{new-aiaa}
\usepackage[utf8]{inputenc}

\usepackage{graphicx}
\usepackage{amsmath}
\usepackage{bm}
\usepackage[version=4]{mhchem}
\usepackage{siunitx}
\usepackage{longtable,tabularx}
\usepackage{booktabs}
\usepackage{svg}
\usepackage{subfig}
\usepackage{tikz}
\usetikzlibrary{angles,quotes,calc,fit}
\usepackage{import}
\usepackage[ruled,vlined]{algorithm2e}
\usepackage{float}
\DeclareSIUnit{\degree}{deg}
\hypersetup{hidelinks}
\DeclareCaptionLabelFormat{rparens}{#2)}
\title{Ground-Wing: A Parametric 3D Wing-in-Ground-Effect Dataset for Benchmarking Neural Surrogate Models}

\author{Andrea Cenzato\footnote{Ph.D. Candidate, Methods for Model-based Development in Computational Engineering}}
\affil{RWTH Aachen University, Aachen, Germany \& BMW Group, Munich, Germany}
\author{Lukas Karch\footnote{Future Computing Lead, Emerging Technologies}}
\author{Andre Luckow\footnote{Department Head, Innovation, Emerging Technologies \& IT Tech Offices}}
\affil{BMW Group, Munich, Germany}
\author{Yoran Pichon\footnote{Head of Aerodynamics CoC}}
\author{Francesco Pasqua\footnote{Aerodynamic Development Engineer}}
\affil{BMW M Motorsport, Munich, Germany}
\author{Abouzar Ghasemi\footnote{HPC Solution Architect}}
\author{Rishikesh Ranade\footnote{Applied Research Lead, Computational Engineering}}
\author{Ira J. S. Shokar\footnote{AI Architect, AI for Science \& Engineering}}
\affil{NVIDIA, Santa Clara, USA}
\author{Julia Kowalski\footnote{Professor, Methods for Model-based Development in Computational Engineering}}
\affil{RWTH Aachen University, Aachen, Germany}

\begin{document}

\maketitle

\begin{abstract}
This paper presents a CFD dataset of a highly parametrized 3D wing operating in ground effect, designed for benchmarking neural surrogate models for aerodynamic prediction, using a wing shape and flow conditions typically encountered in motorsport applications.
The dataset comprises 770 RANS simulations of 194 unique wing configurations, sampled across 27 airfoil profile variants and 4 distances from the ground, spanning variations in wing sweep, chord, taper, angle of attack, and tip twist.
Six neural network architectures representing distinct paradigms---convolutional (DoMINO), graph-based (X-MeshGraphNet), transformer-based (AB-UPT, GeoTransolver, GAOT), and JEPA-based (AeroJEPA)---are evaluated on their ability to predict surface pressure coefficient and wall shear stress distributions, from which integral drag and lift coefficients are recovered. 
The dataset is intended to be used as a challenging benchmark for surrogate CFD models, due to the uniqueness of the ground effect being absent from other wing and airfoil datasets. The dataset is available for download at \url{https://huggingface.co/datasets/andreacenzato/Ground-Wing}.
\end{abstract}


\section{Nomenclature}

{\renewcommand\arraystretch{1.0}
\noindent\begin{longtable*}{@{}l @{\quad=\quad} l@{}}
$C_l$ & lift coefficient \\
$C_d$ & drag coefficient \\
$C_p$ & pressure coefficient \\
$C_{pT}$ & total pressure coefficient \\
$\bm{w_s}$ & wall shear stress vector, \si{\pascal} \\
$\alpha_c$ & centerline angle of attack, \si{\degree} \\
$\theta_t$ & tip twist angle, \si{\degree} \\
$c_c$ & centerline chord length, \si{\milli\meter} \\
$c_t$ & tip chord length, \si{\milli\meter} \\
$\Lambda$ & sweep angle, \si{\degree} \\
$s$ & wing semispan, \si{\milli\meter} \\
$P$ & airfoil profile identifier \\
$\bm{U}$ & velocity vector, \si{\meter\per\second} \\
$\bm{u}$ & nondimensional velocity vector \\
$U_\infty$ & freestream velocity, \si{\meter\per\second} \\
$\rho$ & freestream air density, \si{\kilogram\per\meter\cubed} \\
$Y^+$ & nondimensional wall distance
\end{longtable*}}
\addtocounter{table}{-1}

\section{Introduction}

\lettrine{T}{he} aerodynamic design of front wings for automotive applications relies heavily on computational fluid dynamics (CFD), which, while accurate, demands substantial computational resources and turnaround time. In motorsport applications, for example, fast-paced design cycles require hundreds of geometry iterations within months, making the computational cost of fluid simulations a significant bottleneck. 

The reason for this cost is that a direct analytical solution to the equations which describe the motion of viscous fluids, such as air, has not yet been discovered. The equations in question are known as the Navier--Stokes equations \cite{pope_turbulence}, which CFD software aims to resolve iteratively through finite-volume methods. Even with these strategies, however, simplified modeling of the most complex flow behaviors has to be employed in order to keep the simulation time within reasonable limits. One such simplification, used in this paper, is the usage of the Reynolds-averaged Navier--Stokes (RANS) \cite{rans} equations instead, whose solution provides the time-averaged flow field across the simulation domain, while using models \cite{realisablekepsilon} to calculate the effects of instantaneous turbulence.
In addition, to allow for these methodologies to produce meaningful results, newly designed geometries require extensive preprocessing even before the simulation itself can start. Computer-aided design (CAD) surfaces need to be made perfectly watertight, discretization of the airflow volume needs to be obtained through meshing, and boundary conditions for all relevant surfaces need to be set, all of which require a combination of manpower and computational effort.

Data-driven surrogate models based on neural networks can help alleviate most of these issues. By learning a mapping from geometric inputs directly to flow-field solutions, they can provide predictions of flow variables by avoiding the iterative residual minimization of traditional CFD solvers altogether. They also do not require a mesh to produce a result, as they can operate on simpler representations of the objects of interest, such as simple 3D point clouds, sampled directly from a CAD geometry. This reduces the need for resource-intensive meshing pipelines. By leveraging architectures such as neural operators \cite{neural_operators}, they feature the possibility for predictions on arbitrarily (i.e., distinct from the training setup) dense grids.

However, the deployment of such models hinges on their ability to provide reliable results outside of their training envelope. In order to verify this capability, one option is that of evaluating a trained model on a benchmark dataset related to the desired use case. The model can be trained on part of the benchmark dataset itself or on other, external, data. What matters is that the evaluation set reflects the type of application which the model will be used for in a production environment. Therefore, to expand the usability of surrogate models to a new engineering field, a relevant benchmark dataset is required. One such example is high-performance motorsport, where the complexity of the typical aerodynamic features and the range of geometrical changes between designs make current surrogates unable to provide satisfactory results. In order to reduce this performance gap, exploratory steps can be taken by beginning with evaluating model predictions on single vehicle components, such as the front wing of Le Mans Daytona h (LMDh) prototypes \cite{lmdh_rules}. This component is a crucial element of the vehicle's aerodynamic package, as it greatly affects steering performance, and, subsequently, tire wear and vehicle driveability. This work aims to accomplish this exploratory goal with two main contributions:

\begin{itemize}
    \item Presenting a dataset of RANS solutions of the flow around a 3D downforce-generating wing running in ground effect. The wing geometry is parametrized along five continuous dimensions plus a discrete profile shape parameter, producing a diverse set of configurations that test a model's ability to generalize across planform shape, incidence, and proximity to the ground. Section~\ref{sec:dataset} describes the geometry parametrization, the sampling strategy, and the CFD methodology used to generate the dataset in detail, together with the file formats and fields distributed with each simulation.
    \item Benchmarking six state-of-the-art surrogate models \cite{abupt,domino,xmeshgraphnet,aerojepa,geotransolver,gaot} against the presented dataset. Section~\ref{sec:benchmark} introduces the evaluated neural surrogate architectures, which span the major paradigms in geometric deep learning (convolutional neural operators, graph neural networks, and transformers). It additionally explains the utilized train/validation/test splits, and reports the models' performance on the integral aerodynamic coefficients as well as on the surface fields themselves. Section~\ref{sec:reproducibility} discusses the reproducibility of the benchmarks run on the presented dataset.
\end{itemize}

\section{Related Work}

The development and comparison of neural surrogate models rests on the availability of standardized datasets that capture the complexity of real-world aerodynamic problems. To be useful as a benchmark rather than merely as training data, such a dataset has to supply a geometry family parametrized along documented dimensions, a validated and reproducible CFD setup, surface and volume fields published in open formats, and defined splits on which competing architectures can be compared. A growing number of public CFD datasets meet a subset of these requirements. From them, two main families can be identified: road-car and bluff-body aerodynamics, and wing aerodynamics in free air.

The road-car family spans a wide range of geometric complexity. The earliest entry is the car dataset of Umetani and Bickel \cite{shapenetcar}, commonly referred to as ShapeNet-Car. It contains shapes taken from the car category of the ShapeNet repository \cite{shapenet}, stripped of mirrors, spoilers, and wheels, and simulated with an in-house finite-element solver using a $k$-$\varepsilon$ model, yielding a drag coefficient together with surface pressure and velocity fields. Its shapes are not parametrized, so it supports shape-to-field learning but not a controlled study of parameter generalization, which other bluff-body datasets aim to achieve. AhmedML \cite{ahmedml} contains 500 variants of the Ahmed body generated from 6 geometric parameters, each simulated with a hybrid RANS--large-eddy-simulation (LES) approach in OpenFOAM, while WindsorML \cite{windsorml} contains 355 variants of the Windsor body driven by 7 parameters, computed with a wall-modeled LES on a Cartesian immersed-boundary grid of more than 280M cells. Both distribute time-averaged surface and volume fields, geometry in STL and STEP, and force and moment coefficients. WindsorML additionally publishes deterministic benchmark splits, including geometry- and drag-extrapolation regimes. The remaining road-car datasets use production vehicle geometry. DrivAerNet \cite{drivaernet} morphs the DrivAer fastback along 50 geometric parameters into \num{4000} designs, simulated in OpenFOAM as steady RANS with the $k$-$\omega$ shear-stress transport (SST) model on 16M-cell meshes. DrivAerNet++ \cite{drivaernetpp} extends this to \num{8000} designs across fastback, notchback, and estateback bodies with varying underbody and wheel configurations, described by 26 parameters and 24M-cell meshes, and adds parametric models, part annotations, and point clouds to the flow data. DrivAerML \cite{drivaerml} applies the same hybrid RANS--LES methodology as AhmedML and WindsorML to 500 morphed DrivAer notchback variants on approximately 160M cells. DrivAerStar \cite{drivaerstar} is the largest of the family at \num{12000} STAR-CCM+ simulations spanning three vehicle configurations deformed through 20 CAD parameters, and the only one to model the engine compartment and cooling system with their internal airflow. Finally, SHIFT-SUV \cite{shift_suv} morphs the AeroSUV reference vehicle with a deformation cage sampled by Latin hypercube and computes each variant with a transient detached-eddy simulation.

The wing family is smaller and, with one exception, transonic. AirfRANS \cite{airfrans} is the most widely used reference: \num{1000} two-dimensional steady RANS solutions over NACA four- and five-digit sections at Reynolds numbers between $2 \times 10^6$ and $6 \times 10^6$ and varying angle of attack, computed with OpenFOAM and the Spalart--Allmaras model. Its primary contribution is methodological rather than data-centric, in that it ships four explicitly defined learning tasks, covering abundant and scarce training regimes as well as Reynolds-number and angle-of-attack extrapolation. Three-dimensional wing datasets are instead more recent. SuperWing \cite{superwing} comprises \num{4239} parametrized swept wings and \num{28856} RANS solutions, with the geometry parametrization varying airfoil shape, twist, and dihedral along the span rather than perturbing a single baseline, with each shape simulated across a range of Mach numbers and angles of attack. SHIFT-Wing \cite{shift_wing} instead parametrizes the NASA high-speed Common Research Model through six spanwise airfoil stations and resolves each case with automated mesh adaptation. Emmi-Wing \cite{speedofsound} is the largest, with roughly \num{30000} samples of unique geometry and inflow condition in the highly turbulent transonic regime, computed with a steady compressible OpenFOAM solver and distributed with both surface and volumetric fields.

What none of these datasets contain is a body operating in close proximity to a wall. Every one of them places its geometry either in free air or, for the road cars, at a fixed ride height that is not itself a design variable, so ground clearance never appears as a dimension along which a surrogate model is asked to generalize. Ground effect introduces pronounced nonlinearities in the pressure field due to the confined flow between the wing and the ground plane, with phenomena such as enhanced Venturi acceleration and potential flow choking at low clearances, and it is precisely this regime that is of greatest interest to front-wing design in motorsport. The dataset presented in Section~\ref{sec:dataset} is built to fill that gap, varying both the wing shape and its distance from a moving ground plane, and pairing the simulations with the splits and metrics needed to use them as a benchmark.

\section{Dataset Presentation}
\label{sec:dataset}

\subsection{Geometry Parametrization}
\label{sec:parametrization}

\begin{figure}[hbt!]
    \centering
    \subfloat[Rendering of a wing geometry sample.]{\label{fig:wing_cad}
        \includegraphics[width=0.48\linewidth]{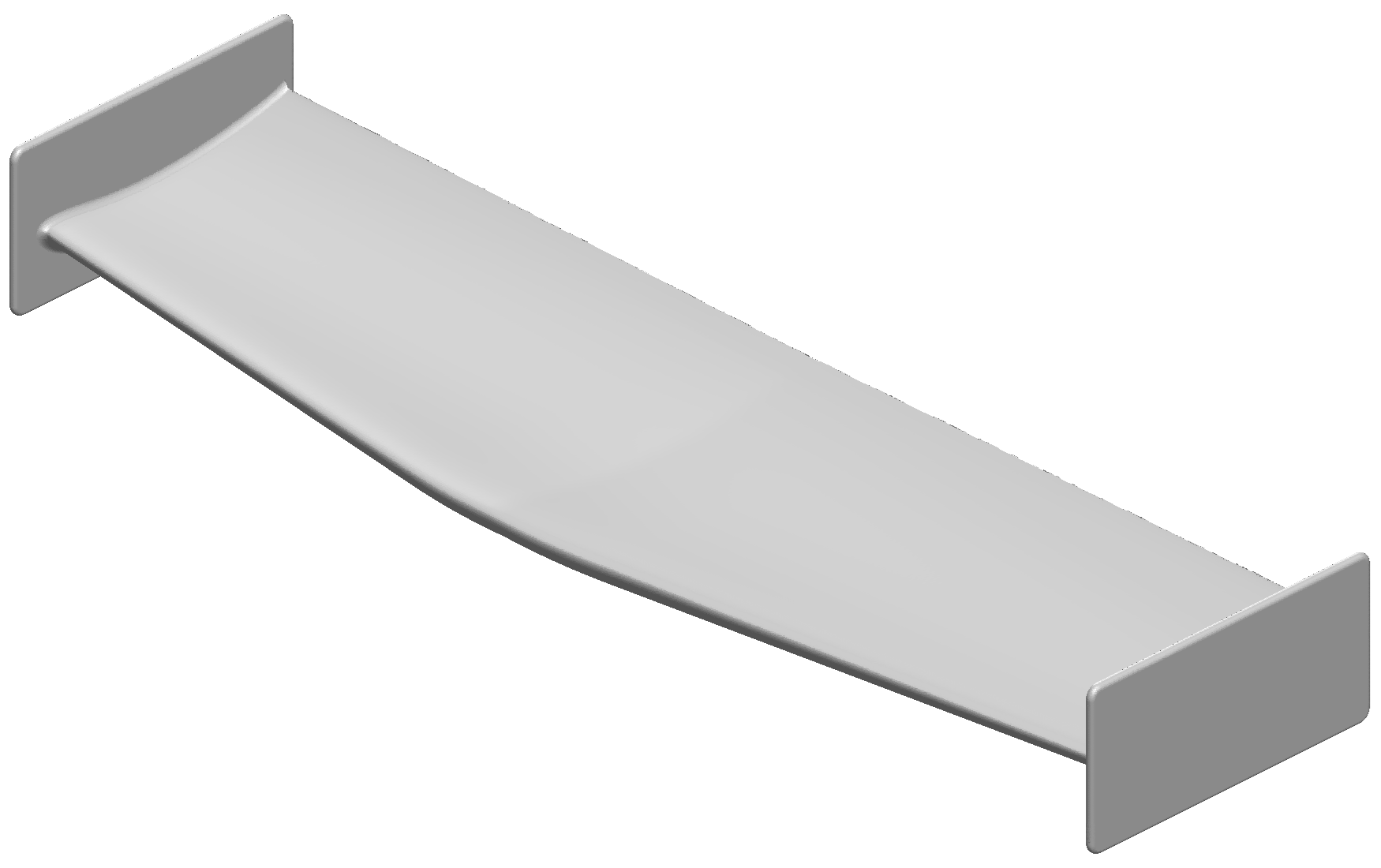}}
    \hfill
    \subfloat[Top view of the ground plane, with the slip floor in purple, the moving belt in blue, and the wing in green.]{\label{fig:wt_floor}
        \includegraphics[width=0.48\linewidth]{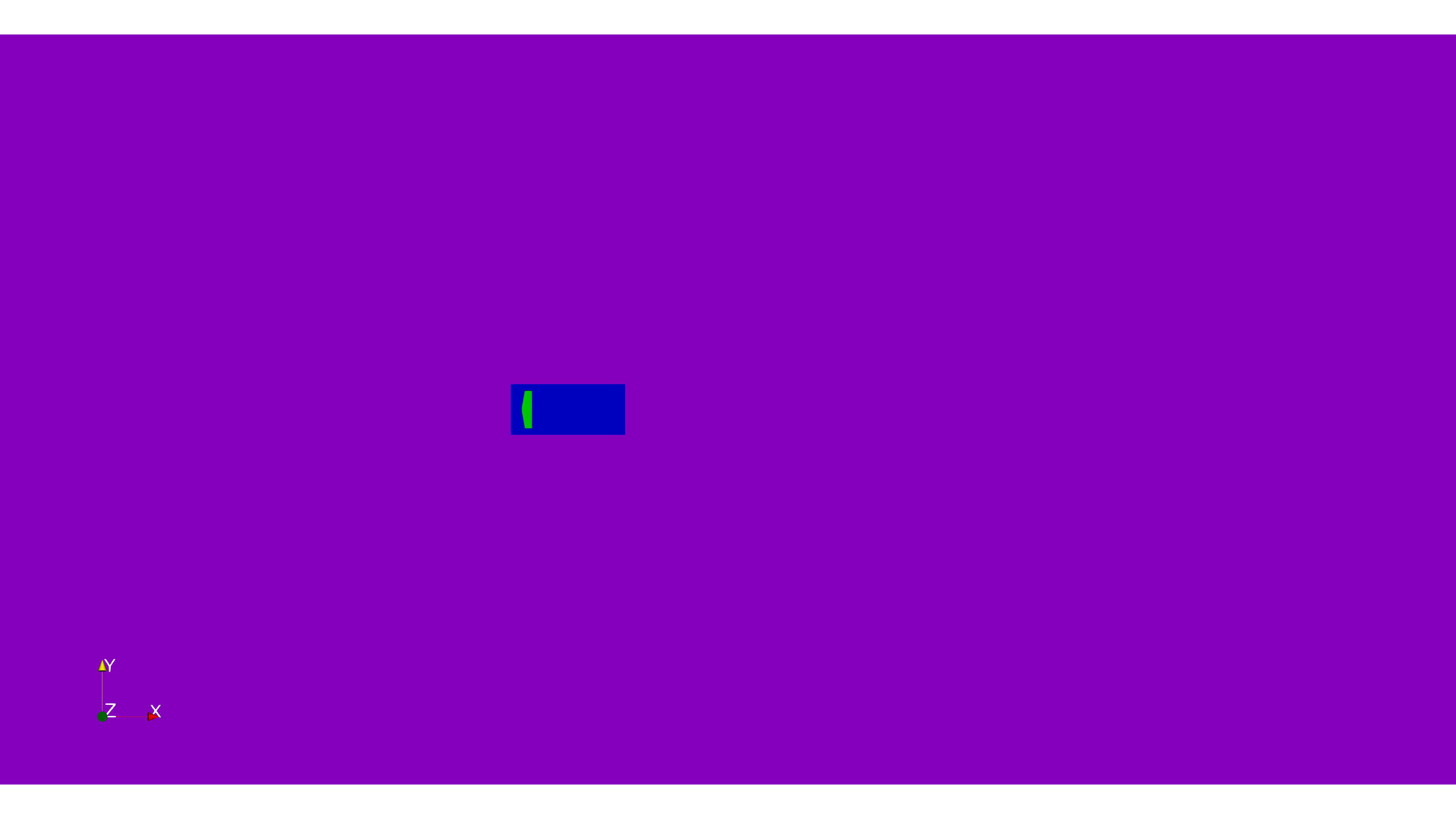}}
    \caption{Wing geometry sample and its placement on the wind-tunnel ground plane.}
\end{figure}

The geometry is a single-element wing with rectangular endplates, shown in Fig.~\ref{fig:wing_cad}, designed to loosely resemble and occupy the same design space as the front splitter of an LMDh prototype race car \cite{lmdh_rules}. This choice was guided by the desire to maintain real-world relevance. The geometry is designed to achieve two competing characteristics: complexity representative of modern race cars, and simplicity required for parametric shape generation. A single element was retained because the additional complexity introduced by the interaction between multiple closely coupled profiles is beyond the scope of this dataset. Endplates, conversely, were included on every geometry, since wings used in motorsport almost never operate without them or without an equivalent tip device, given the performance benefit they provide \cite{endplate1,endplate2}.

The CAD model was built parametrically in Dassault's CATIA, with the design space defined by classifying parameters into variable inputs that drive aerodynamic performance and fixed values that maintain relevance to the specific application and CAD model stability (Table~\ref{tab:params}). Beyond the parameters of Table~\ref{tab:params}, the CAD model exposes a set of endplate and fillet dimensions that are held fixed across all geometries, as their influence on the aerodynamic coefficients was found to be secondary to that of the primary parameters, and keeping them fixed reduces the dimensionality of the design space without sacrificing geometric realism. The endplate sizing follows recommendations from the ground-effect literature \cite{endplate1,endplate2}, while the fillet radii serve to smooth the overall shape of the wing and were sized to avoid local flow separation. The values of these parameters, the reasoning behind each of them, and additional information about the geometry definition are described in Appendix~\ref{sec:appendix_geometry}.

\begin{figure}[hbt!]
    \centering
    \subfloat[Front view of the wing between its two endplates, showing the ride height (equal to the endplate clearance) and the resulting profile ground clearance.]{\label{fig:geom_clearance}
    \begin{minipage}[b]{0.48\linewidth}
    \centering
    \begin{tikzpicture}[x=0.60cm, y=0.72cm, font=\footnotesize]
        \def\halfSpan{5}
        \def\plateW{0.2}
        \def\rideHeight{1}
        \def\plateTop{4.8}
        \def\profileThick{0.45}
        \pgfmathsetmacro{\profileClearance}{(\rideHeight+\plateTop)/2-\profileThick/2}

        \coordinate (Origin) at (-1.3,0);

        \draw[->, thick, loosely dash dot] (Origin) -- ({2*\halfSpan+0.5},0) node[right] {$y$};
        \draw[->, thick, loosely dash dot] (Origin) -- ++(0,4.2) node[above] {$z$};
        \node[below] at ({\halfSpan},0) {Ground};

        \draw[thick, fill=gray!20, rounded corners=3pt]
          (0,\rideHeight) rectangle (\plateW,\plateTop);
        \draw[thick, fill=gray!20, rounded corners=3pt]
          ({2*\halfSpan-\plateW},\rideHeight) rectangle ({2*\halfSpan},\plateTop);

        \draw[thick, fill=gray!20]
          (\plateW,\profileClearance) rectangle ({2*\halfSpan-\plateW},{\profileClearance+\profileThick});

        \draw[dashed] (\plateW,\rideHeight) -- ({\halfSpan-0.4},\rideHeight);

        \draw[<->, thick] ({\halfSpan-0.4},0) -- ({\halfSpan-0.4},\rideHeight);
        \node[anchor=south east, align=right, inner sep=2pt] at ({\halfSpan-0.45},{\rideHeight+0.08}) {Ride height $=$ \\ Endplate clearance};
        \draw[<->, thick] ({\halfSpan+1},0) -- ({\halfSpan+1},\profileClearance);
        \node[anchor=west, align=left] at ({\halfSpan+1.2},{\profileClearance/2}) {Profile \\ clearance};
    \end{tikzpicture}
    \end{minipage}
    }
    \hfill
    \subfloat[Visualization of centerline angle of attack $\alpha_c$ and tip twist $\theta_t$.]{\label{fig:geom_angles}
    \begin{minipage}[b]{0.48\linewidth}
    \centering
    {\footnotesize\def\svgwidth{\linewidth}
\begingroup%
  \makeatletter%
  \providecommand\color[2][]{%
    \errmessage{(Inkscape) Color is used for the text in Inkscape, but the package 'color.sty' is not loaded}%
    \renewcommand\color[2][]{}%
  }%
  \providecommand\transparent[1]{%
    \errmessage{(Inkscape) Transparency is used (non-zero) for the text in Inkscape, but the package 'transparent.sty' is not loaded}%
    \renewcommand\transparent[1]{}%
  }%
  \providecommand\rotatebox[2]{#2}%
  \newcommand*\fsize{\dimexpr\f@size pt\relax}%
  \newcommand*\lineheight[1]{\fontsize{\fsize}{#1\fsize}\selectfont}%
  \ifx\svgwidth\undefined%
    \setlength{\unitlength}{349.90125625bp}%
    \ifx\svgscale\undefined%
      \relax%
    \else%
      \setlength{\unitlength}{\unitlength * \real{\svgscale}}%
    \fi%
  \else%
    \setlength{\unitlength}{\svgwidth}%
  \fi%
  \global\let\svgwidth\undefined%
  \global\let\svgscale\undefined%
  \makeatother%
  \begin{picture}(1,0.42308193)%
    \lineheight{1}%
    \setlength\tabcolsep{0pt}%
    \put(0,0){\includegraphics[width=\unitlength,page=1]{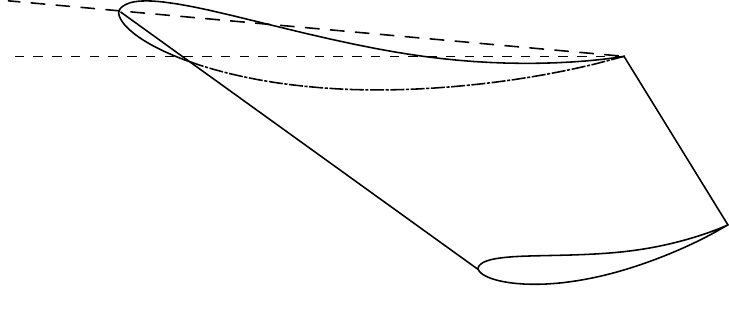}}%
    \put(-0.0023793,0.37211505){\color[rgb]{0,0,0}\makebox(0,0)[lt]{\lineheight{1.25}\smash{\begin{tabular}[t]{l}$\alpha_c$\end{tabular}}}}%
    \put(0.42631305,0.12871164){\color[rgb]{0,0,0}\makebox(0,0)[lt]{\lineheight{1.25}\smash{\begin{tabular}[t]{l}$\alpha_c$\end{tabular}}}}%
    \put(0.35129189,0.07911173){\color[rgb]{0,0,0}\makebox(0,0)[lt]{\lineheight{1.25}\smash{\begin{tabular}[t]{l}$\theta_t$\end{tabular}}}}%
    \put(0,0){\includegraphics[width=\unitlength,page=2]{profile.pdf}}%
  \end{picture}%
\endgroup%
}
    \end{minipage}
    }\\
    \subfloat[Legality-box constraint on the wing planform in the $xy$ plane.]{\label{fig:geom_box}
    \begin{minipage}[b]{0.75\linewidth}
    \centering
    \begin{tikzpicture}[x=0.78cm, y=0.78cm, font=\footnotesize]
        \def\rootChord{4}
        \def\tipChord{2}
        \def\span{10}
        \def\sweepAngle{20}
        \def\maxX{7}

        \coordinate (O) at (0,-1);
        \coordinate (X) at (0,8);
        \coordinate (Y) at (-10,0);

        \coordinate (LEroot) at (0,0.5);
        \coordinate (LEtip) at ($(LEroot) + (-{\span * cos(\sweepAngle)}, {\span * sin(\sweepAngle)})$);
        \coordinate (LEoff) at ($(LEroot) + (-3,0)$);

        \coordinate (TEroot) at ($(LEroot) + (0,\rootChord)$);
        \coordinate (TEtip) at ($(LEtip) + (0,\tipChord)$);

        \draw[thick, fill=gray!20]
          (LEroot) -- (LEtip) -- (TEtip) -- (TEroot) -- cycle;

        \draw[->, thick, loosely dash dot] (O) -- (X) node[above right] {$x$};

        \def\offset{0.3}

        \draw[dashed, red, thick] (\offset,\maxX) -- node[above] {Max $x$} ++($(Y) + (-\offset,0)$);
        \draw[dashed, red, thick] (\offset,0) -- node[below] {Min $x$} ++($(Y) + (-\offset,0)$);

        \draw[very thin] (LEroot) -- ++(-2,0);
        \draw pic["$\Lambda$", draw=black, angle radius=9.5mm, angle eccentricity=1.3]
          {angle=LEtip--LEroot--LEoff};

        \draw[<->, thick]
          ($(LEtip) - (\offset,0)$) -- node[left] {$c_t$} ($(TEtip) - (\offset,0)$);
        \draw[<->, thick]
          ($(LEroot) + (\offset,0)$) -- node[right] {$c_c$} ($(TEroot) + (\offset,0)$);
        \draw[<->, thick]
        	($(Y) + (-\offset,0)$) -- node[left] {\SI{700}{\milli\meter}} ++(0,\maxX);
        \draw[<->, thick]
        	($(TEtip) + (0,\offset)$) -- node[above] {$s$} ++({\span * cos(\sweepAngle)}, 0);
    \end{tikzpicture}
    \end{minipage}
    }
    \caption{Geometric constraints and angular parameters of the wing planform, and its ride-height and ground-clearance conventions.}
\end{figure}

The wing span and the minimum profile ground clearance are taken from the corresponding dimensions of the reference race car's front wing, where the span is dictated by the vehicle's bodywork. Variations in ground clearance are achieved through a ride-height sweep described in Section~\ref{sec:cfd}. Throughout this paper, the ride height denotes the clearance between the ground and the lowest point of the endplate, which is the lowest point of the whole assembly on every geometry (Fig.~\ref{fig:geom_clearance}). Extending the endplates as far downwards as possible follows the aforementioned design guidelines, and compensates for the reference vehicle controlling the flow underneath the wing with other aerodynamic devices rather than with endplates. The bounds on the two chord lengths and on the sweep angle follow from the regulatory legality box, which limits how far forward the wing may reach, and from the front wheel arches, which limit it rearwards. Figure~\ref{fig:geom_box} illustrates the legality-box constraint on the planform, discussed further in Section~\ref{sec:sampling}. The lower chord bounds were instead selected as a reasonable minimum size for the application. Finally, the angle of attack (AoA) and tip twist ranges are chosen such that every used airfoil profile operates without stall or large-scale separation along the entire span, avoiding flow features that are generally undesired in front-wing design. Figure~\ref{fig:geom_angles} visualizes the centerline angle of attack $\alpha_c$ and tip twist $\theta_t$ definitions.

Additionally, two further simplifications were introduced to keep the CAD model stable and the parameter count tractable, at a modest cost in geometric realism:
\begin{itemize}
    \item \textbf{Constant profile and ground clearance along span}: both the airfoil profile and the minimum ground clearance are held constant along the span, precluding a raised center section or a spanwise blend between two profiles, in exchange for a much lower-dimensional design space. This is achieved by having the lowest point of the airfoil section be tangent to a horizontal plane across the entire span.
    \item \textbf{Linear sweep and twist}: the sweep angle and tip twist vary at a constant rate across the span, rather than allowing a curved leading edge or abrupt local twist changes, so that the planform is fully described by a single sweep and twist value each rather than a spanwise distribution.
\end{itemize}

\begin{table}[hbt!]
\caption{\label{tab:params} Wing geometry parameters: variable design-space ranges (left), fixed values (center), and the ride heights simulated for every geometry (right).}
\centering
\small
\begin{tabular}{lccclccc}
\toprule
\multicolumn{4}{c}{Variable} & \multicolumn{2}{c}{Fixed} & \multicolumn{2}{c}{Ride heights} \\
\cmidrule(lr){1-4}\cmidrule(lr){5-6}\cmidrule(lr){7-8}
Parameter & Symbol & Min & Max & Parameter & Value & Endplate & Profile \\
\midrule
Profile & $P$ & --- & --- & Wing span & \SI{2000}{\milli\meter} & \SI{30}{\milli\meter} & \SI{87}{\milli\meter} \\
Centerline AoA & $\alpha_c$ & \SI{-10}{\degree} & \SI{5}{\degree} & & & \SI{45}{\milli\meter} & \SI{102}{\milli\meter} \\
Tip twist & $\theta_t$ & \SI{-10}{\degree} & \SI{10}{\degree} & & & \SI{60}{\milli\meter} & \SI{117}{\milli\meter} \\
Centerline chord & $c_c$ & \SI{550}{\milli\meter} & \SI{700}{\milli\meter} & & & \SI{75}{\milli\meter} & \SI{132}{\milli\meter} \\
Tip chord & $c_t$ & \SI{350}{\milli\meter} & \SI{700}{\milli\meter} & & & & \\
Sweep angle & $\Lambda$ & \SI{-8.5}{\degree} & \SI{20}{\degree} & & & & \\
\bottomrule
\end{tabular}
\end{table}

The profile parameter $P$ determines which airfoil profile is used to construct the main wing body. It is drawn from a pool of 27 total options, constructed as follows:

\begin{itemize}
    \item \textbf{Base set}: 12 publicly available \cite{airfoiltools,uiuc_airfoils} high-lift shapes commonly used in motorsport applications, or developed specifically for motorsport use \cite{mshd,benzing}, complemented by three standard NACA sections. The non-NACA sections all share high camber and comparatively small thickness. These profiles form the main corpus of the dataset.
    \item \textbf{Interpolated set}: 15 additional sections, used to evaluate model performance on profile shapes unseen during training:
    \begin{itemize}
        \item 2 high-lift profiles (S1223RTL, S3021), retrieved from \cite{airfoiltools}.
        \item 13 pairwise interpolated blends between BE122-125 and each of the other base-set sections and the 2 additional high-lift profiles listed above, obtained as simple coordinate averages with XFOIL \cite{xfoil}.
    \end{itemize}
\end{itemize}

Appendix~\ref{sec:appendix_profiles} names all 27 variants and identifies which of them belong to the base set, and the same appendix describes how the profiles were selected and how the interpolated ones were generated.

\subsection{Sampling Strategy}
\label{sec:sampling}

To determine which geometries to generate from the available parameter combinations, two objectives were pursued:
\begin{enumerate}
    \item The samples must cover the five-dimensional continuous design space of Table~\ref{tab:params} without leaving blind spots, so that no region of the space is systematically mispredicted for lack of training data. 
    \item Samples should be denser where the aerodynamic response to parameter variations changes most rapidly, as these regions are the most complex areas of the design space for the model to learn. Concentrating samples where the informativeness of an additional sample is highest is common practice in data-centric machine learning (ML) \cite{equalsamples}, where it is normally used to select or weight samples of an existing dataset. The same reasoning is applied here one step earlier, to decide which simulations to even run.
\end{enumerate}

The first objective is addressed by filling the design space with a Sobol sequence \cite{sobol}, a sampling strategy that fills a unit hypercube by aiming to distribute samples as evenly as possible across all dimensions, avoiding the clusters and voids produced by random sampling at any finite sample count. An additional feasibility filter is applied to every candidate generated by the Sobol sequence, to eliminate combinations of parameters that, while each individually lying inside the boundaries of Table~\ref{tab:params}, produce geometries that do not respect the constraints of Section~\ref{sec:parametrization}. The filter is implemented as two families of constraints. The first bounds the sum of centerline angle of attack and tip twist, to ensure that no spanwise station of the wing operates outside the incidence range in which all profiles of the pool remain free of separation and stall. The second keeps the planform within the legality box of Section~\ref{sec:parametrization}. Since the semispan is fixed, sweeping the wing displaces its tip longitudinally with respect to the centerline, so a large sweep angle combined with a large chord pushes the geometry beyond the allowed margins even when both parameters are individually admissible. Candidates violating any of these conditions are discarded before any selection takes place. Appendix~\ref{sec:appendix_sampling} provides a further explanation of the application of the Sobol sequence, and of the constraint equations used in the feasibility filter.

The second objective is addressed by not simulating the Sobol points directly. Instead, a Sobol set large enough that at least three times the target number of geometries survives the feasibility filter is generated, and the surviving candidates are then reduced to the final sample count by a weighted selection driven by a sensitivity function
\begin{equation}
    S(p) = \left| \frac{\partial C_l}{\partial p} \right| + \left| \frac{\partial C_d}{\partial p} \right|,
    \label{eq:sensitivity}
\end{equation}
defined separately for every continuous parameter $p$ of Table~\ref{tab:params}. The rationale for this particular form is that the predicted quantities the surrogate models are ultimately judged on are the integral coefficients, so a region of the design space is considered informative exactly when a small geometric perturbation there produces a large change in $C_l$ or $C_d$. The usage of the absolute value ensures that high gradients in any direction produce a refinement of the sampling region.

The derivatives in Eq.~\eqref{eq:sensitivity} are obtained by CFD. First, each parameter is independently swept over its full range about a fixed baseline geometry, with all other parameters held constant and with the same solver setup used for the dataset itself (Section~\ref{sec:cfd}). From the results of the simulations, cubic splines are fitted through the resulting $C_l(p)$ and $C_d(p)$ curves for each parameter, and are differentiated analytically. This univariate construction deliberately ignores interactions between parameters: resolving the full five-dimensional response surface would have required a number of simulations comparable to the dataset itself, which would defeat the purpose of using it to decide where to sample. No sensitivity function is defined for the discrete profile parameter $P$, since it does not correspond to a continuously variable dimension and a derivative with respect to it is meaningless. Profiles are instead assigned to the selected parameter sets through a separate Sobol distribution.

In the weighted selection, the sensitivity weight of a candidate sample is the product of the $S(p)$ values of its parameters, so that a sample is rated informative only if it is informative in every dimension at once. That weight is combined with a coverage term, recomputed after every draw, which favors candidates far from the already-selected set. The two weights are normalized and given equal importance. These terms pull in opposite directions by construction, and weighting them equally is what keeps the sensitivity weighting from collapsing the sample set onto the most sensitive corners of the design space.

\begin{figure}[hbt!]
    \centering
    \includesvg[width=\linewidth]{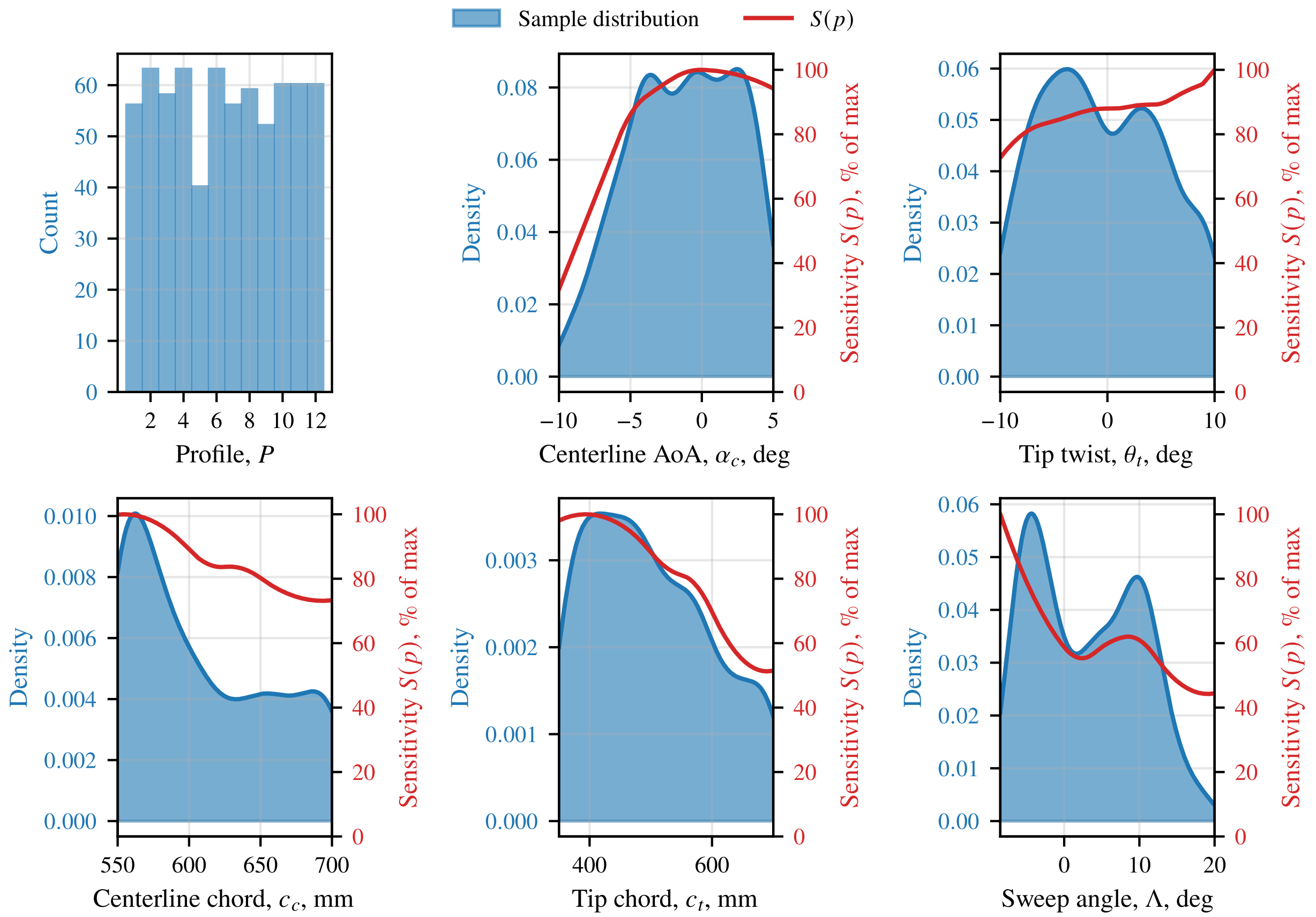}
    \caption{Sampling density of each geometric parameter across the 174 base-set geometries (blue, left axis), with the sensitivity function $S(p)$ of Eq.~\eqref{eq:sensitivity} overlaid as a percentage of its maximum over each parameter range (red, right axis). The discrete profile parameter $P$ is shown as counts.}
    \label{fig:param_dist}
\end{figure}

Applying this procedure yields the 194 unique wing geometries that constitute the dataset: 174 from the base set of profiles, and 20 built from the interpolated set of airfoil variants (Section~\ref{sec:parametrization}). Figure~\ref{fig:param_dist} shows the resulting sampling density of each geometric parameter over the geometries built with the base set of airfoils, together with the sensitivity function of Eq.~\eqref{eq:sensitivity}, normalized to its maximum over each parameter range. The distributions do not follow $S(p)$ exactly, as they are not meant to: the coverage term and the feasibility filter both flatten them, the latter visibly so at the edges of the $\alpha_c$ and $c_t$ ranges, where large parts of the hypercube are infeasible. The imprint of the sensitivity weighting is nonetheless clear, most notably in the peaks of the $\theta_t$ and $\Lambda$ distributions and in the higher density at small $c_c$ and $c_t$, where $S(p)$ is largest.

Appendix~\ref{sec:appendix_sampling} describes the setup used to obtain the sensitivity functions in detail, along with a deeper breakdown of the weighted selection process. Appendix~\ref{sec:appendix_interaction} instead inspects the magnitude of the error introduced by not considering cross-parameter interactions in the computation of the sensitivity functions.

\subsection{CFD Methodology}
\label{sec:cfd}

All simulations are performed using Siemens STAR-CCM+ with an incompressible RANS approach, a realizable $k$-$\varepsilon$ turbulence model \cite{realisablekepsilon}, and a runtime of \num{1000} iterations. The freestream velocity is $U_\infty = \SI{50}{\meter\per\second}$, parallel to the wing's centerline chord direction, with $\rho = \SI{1.18858}{\kilogram\per\meter\cubed}$. The computational domain measures $\SI{30}{\meter}\times\SI{40}{\meter}\times\SI{80}{\meter}$ (height~$\times$~width~$\times$~length). A slip condition is applied to the bottom surface of the simulation domain, except for a $\SI{3}{\meter}\times\SI{10}{\meter}$ area beneath the wing, which moves at $U_\infty$, and where a no-slip condition is applied (Fig.~\ref{fig:wt_floor}).

\begin{figure}[hbt!]
    \centering
    \includegraphics[width=0.85\linewidth]{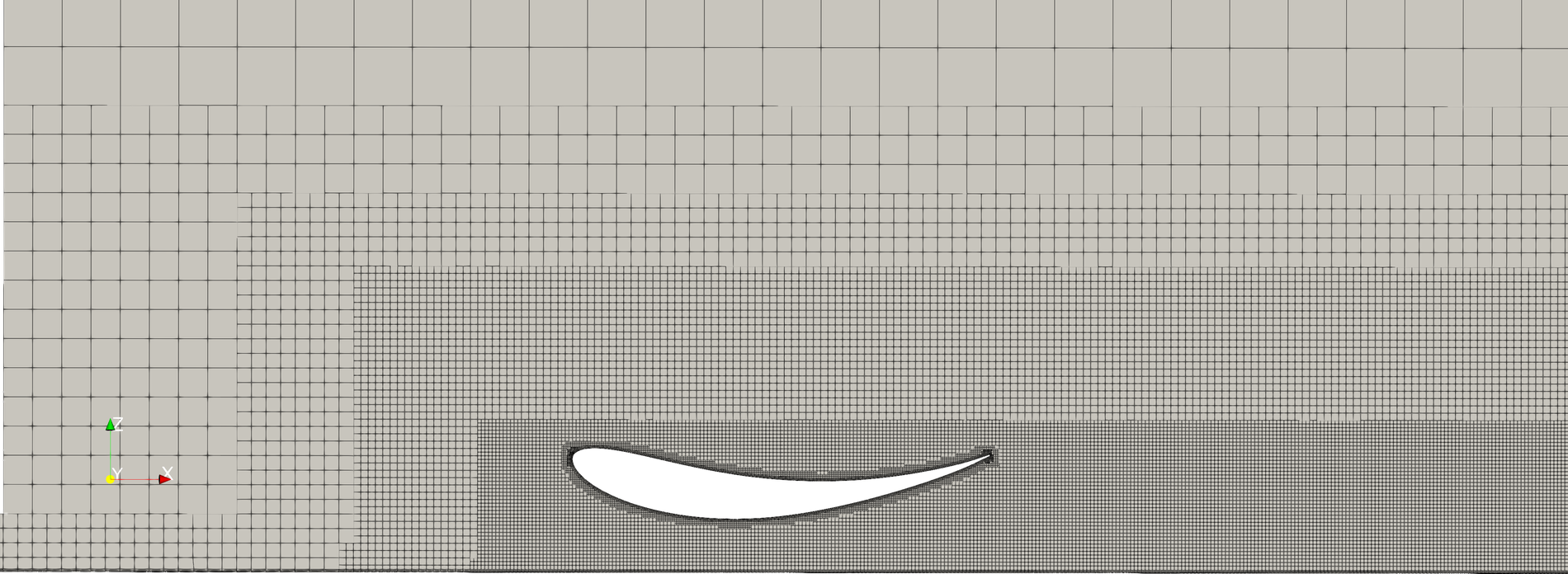}
    \caption{Mesh section at the wing centerline plane.}
    \label{fig:mesh}
\end{figure}

\begin{figure}[hbt!]
    \centering
    \subfloat[$C_l$ convergence]{\label{fig:mesh_study_cl_conv}
        \includesvg[width=0.48\linewidth]{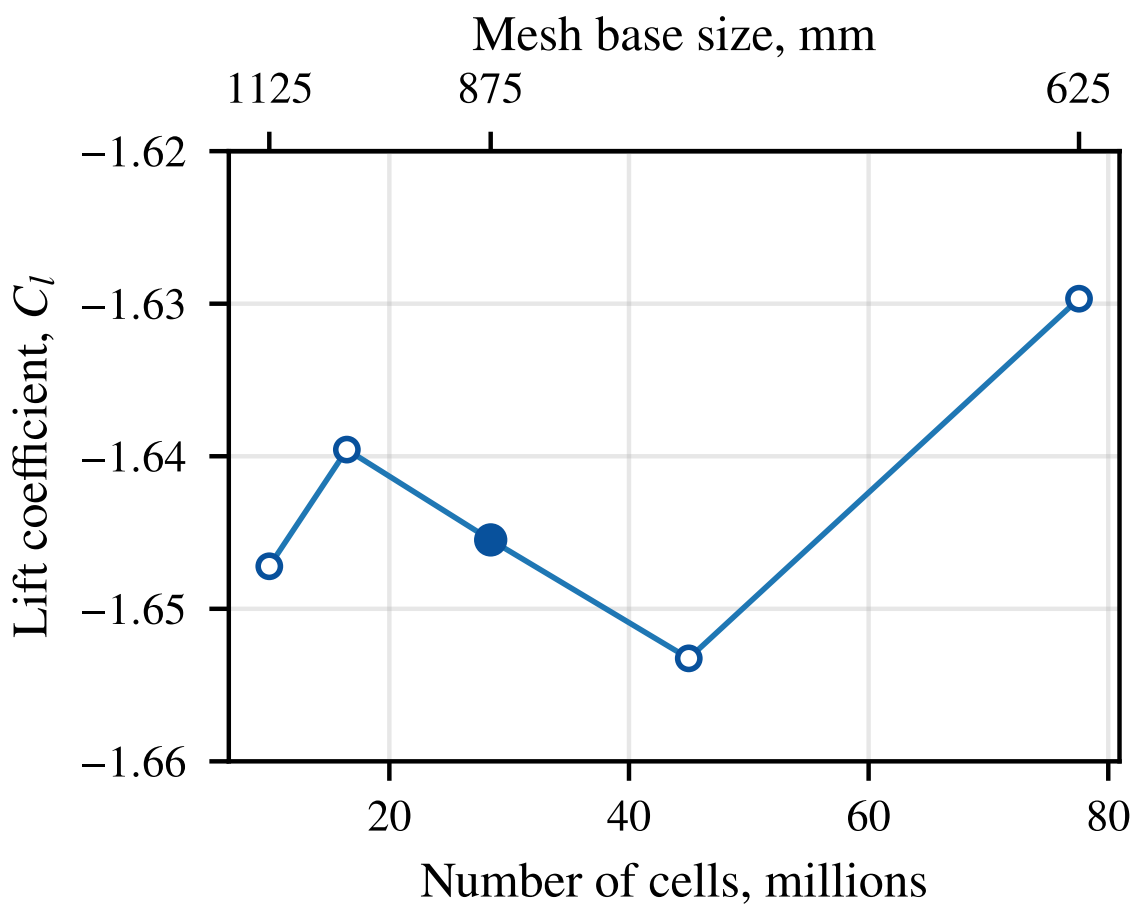}}
    \hfill
    \subfloat[$C_l$ percentage error]{\label{fig:mesh_study_cl_err}
        \includesvg[width=0.48\linewidth]{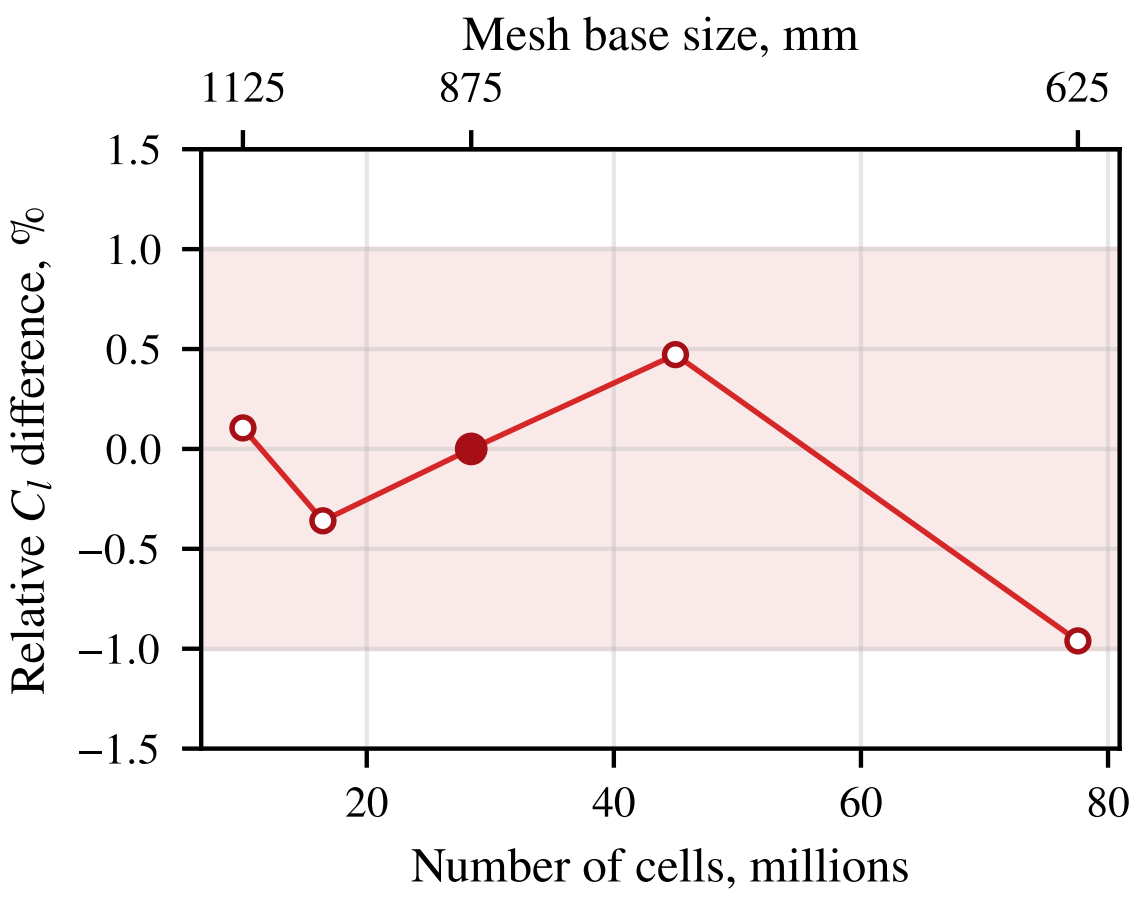}}\\
    \subfloat[$C_d$ convergence]{\label{fig:mesh_study_cd_conv}
        \includesvg[width=0.48\linewidth]{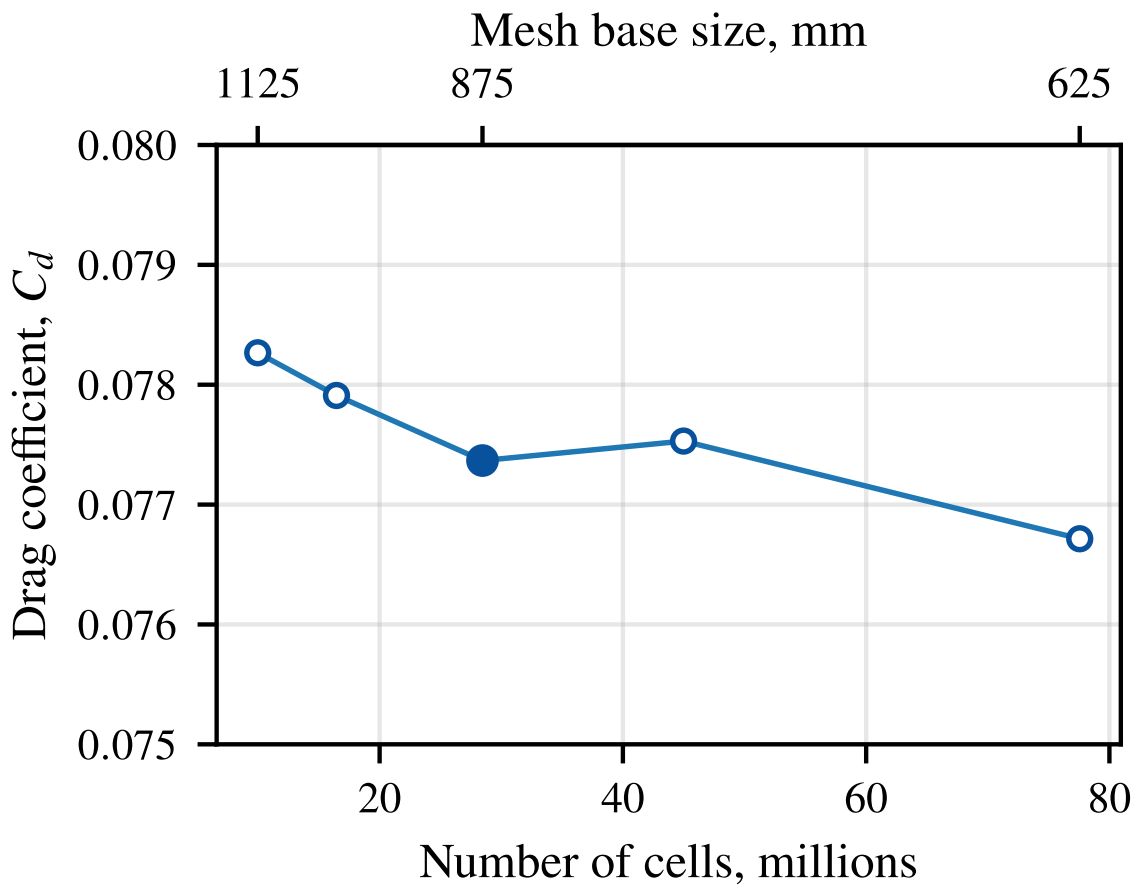}}
    \hfill
    \subfloat[$C_d$ percentage error]{\label{fig:mesh_study_cd_err}
        \includesvg[width=0.48\linewidth]{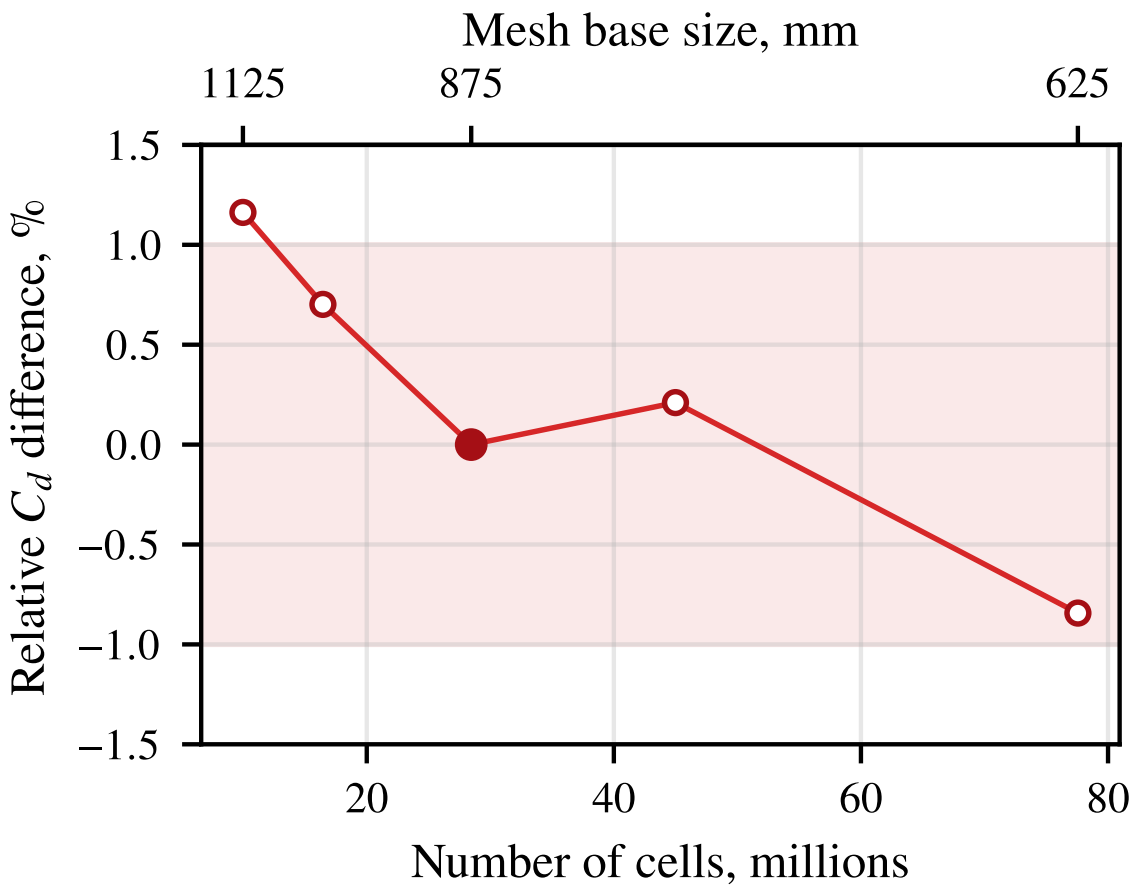}}
    \caption{Aerodynamic coefficient sensitivity to mesh base size, relative to the 28M-cell configuration used for the dataset.}
    \label{fig:mesh_study}
\end{figure}

The hexahedral mesh uses a base size of \SI{875}{\milli\meter} with 9 levels of refinement, halving the cell edge length at each level and yielding a minimum surface cell size of \SI{1.5}{\milli\meter} and approximately 28 million cells per simulation (Fig.~\ref{fig:mesh}). A dedicated mesh independence study, varying the base size between \SI{625}{\milli\meter} and \SI{1125}{\milli\meter} (Fig.~\ref{fig:mesh_study}), showed that the chosen 28M-cell mesh is within 1\% of a 77M-cell reference mesh for both $C_l$ and $C_d$, a compromise favoring the large number of simulations required for the dataset over marginal per-case accuracy. A separate study increasing the number of prism-layer inflation cells (from 5 to 20, at fixed total height) to drive the wall-adjacent $Y^+$ from the range of 30--100 down to $Y^+\simeq 1$ changed both coefficients by only 0.6\%, confirming that wall-function modeling of the boundary layer is adequate at the chosen resolution. With the selected settings, each simulation required around \SI{560}{\second} on 320 CPU cores. For the complete dataset, this translates to $\simeq\SI{120}{\hour}$ of real compute time, and $\simeq\num{38300}$ core hours.

The ride heights of Table~\ref{tab:params} span the range of ground clearances relevant to race-car setup work. Each is reproduced as a rigid vertical translation of the wing relative to the bottom surface of the simulation domain. On the reference vehicle, that distance is set by the suspension ride height, which is \SI{10}{\milli\meter} lower than the values quoted here because of the offset between the suspension reference and the endplate's lower edge. Only flat setups, i.e., equal front and rear ride heights, are included in the dataset, as with no other body present in the flow field, a change in chassis rake angle is equivalent to a change in angle of attack up to the rotation of the endplates, and would therefore duplicate information already carried by $\alpha_c$. Appendix~\ref{sec:appendix_geometry} details this equivalence.

\begin{figure}[hbt!]
    \centering
    \includesvg[width=0.75\linewidth]{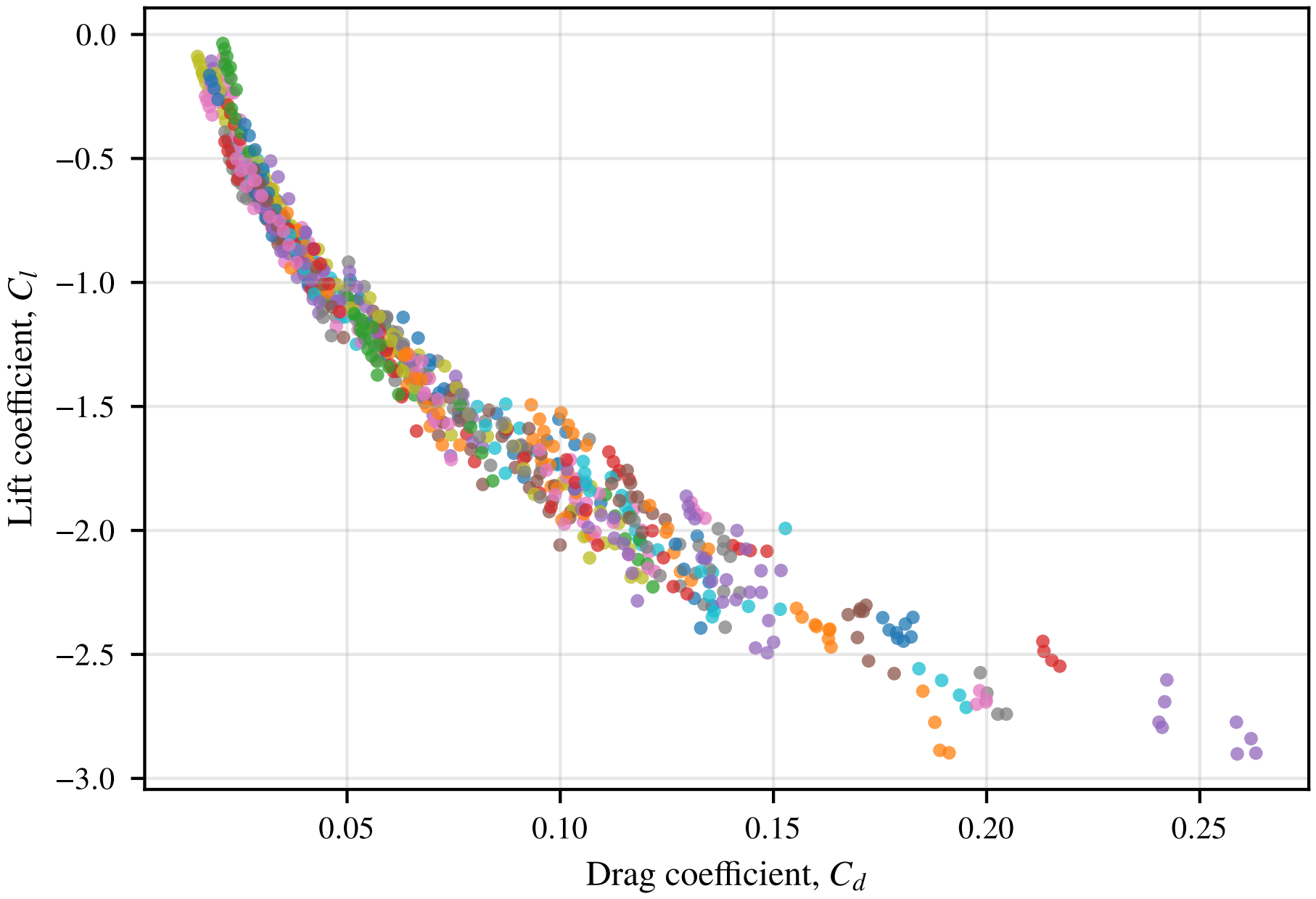}
    \caption{$C_d$ vs.\ $C_l$ for all 770 simulated samples, with each unique wing geometry distinguished by color.}
    \label{fig:cd_cl_all}
\end{figure}

Figure~\ref{fig:cd_cl_all} shows the aerodynamic envelope covered by the dataset across all 770 simulated samples, with each unique wing geometry assigned a distinct color across its simulated ride heights.

Ground-effect wings are additionally susceptible to a nonlinear ``choking'' phenomenon at very low clearances, in which the flow beneath the wing separates and downforce collapses \cite{inverted_wing}. A dedicated ride-height sweep, reported in Appendix~\ref{sec:appendix_ride_height}, confirms that all four ride heights of Table~\ref{tab:params} lie within the linear, well-behaved operating region.

\subsection{Simulation Output and File Format}
\label{sec:fileformat}

Each simulation is identified by one four-digit \textit{run ID}, which identifies a unique geometry, and one four-digit \textit{ride height}, which identifies the distance from the ground of a specific sample. The resulting name convention is the following: sample 0123 at a distance from the ground of \SI{75}{\milli\meter} will be named \textit{run\_0123\_7575}.

For each simulation, three files are distributed: a surface-only \texttt{.stl} geometry (no field data, provided for reference and for direct use by geometry-only models), a surface \texttt{.vtp} file with the cell-centered fields $C_p$, $C_{pT}$, and $\bm{w_s}$ (in \si{\pascal}), and a volumetric \texttt{.vtu} file with point data for $C_p$, $C_{pT}$, $\bm{U}$, and $\bm{u}$. Additionally, a machine-readable database, \texttt{run\_db.csv}, provides information about each run's geometric parameters and integrated force coefficients. The surface meshes contain approximately 1.2M surface cells, while the volume meshes contain approximately 28M cells.

\section{Neural Surrogate Benchmarking}
\label{sec:benchmark}

The benchmark phase of this paper focuses on testing the performance of current state-of-the-art surrogate models on the task of predicting the surface variables $C_p$ and $\bm{w_s}$. The models used in the benchmark were selected due to their stated prediction performance, their ability to handle large, multimillion-element meshes, and their distinct architectural paradigms.

Section~\ref{subsec:splits} defines the train/validation/test splits on which the models are trained and evaluated. Section~\ref{subsec:models} describes the main concept behind each model and how it was configured, and Section~\ref{subsec:metrics} defines the error metrics. The results follow in Sections~\ref{subsec:results} and~\ref{subsec:ood}, with Section~\ref{subsec:delta} assessing whether the models rank geometries correctly and Section~\ref{subsec:fields} providing a qualitative comparison of the local surface pressure error.

\subsection{Dataset Splits}
\label{subsec:splits}

Three separate dataset splits are utilized to evaluate the models' capabilities on samples lying both inside and outside of the training envelope. 

\subsubsection{Stratified Split}
The collection of wings generated using the base set of airfoil profiles is split into training, validation, and test subsets using iterative stratification \cite{iterstrat}. This splitting strategy is designed for datasets characterized by multiple continuous and categorical parameters, and aims to preserve the marginal distribution of every parameter across all subsets. The split acts on the 174 unique geometries rather than on the 690 individual simulations, and the available ride heights of a geometry are reattached to whichever subset that geometry was assigned to. No geometry is therefore shared between training, validation, and test at a different ride height, which would otherwise let a model recognize a test shape it had already been trained on. To build the label vector that stratification operates on, the five continuous parameters of Table~\ref{tab:params} are discretized into 15 uniform bins each and one-hot encoded together with the categorical profile parameter $P$, turning every geometry into a multilabel indicator vector. The resulting set is then split twice, first 80/20 and then the 20\% half again into equal parts. The procedure is stated in full in Appendix~\ref{sec:appendix_splits}.

This is the split whose test set the results of Section~\ref{subsec:results} refer to. Six cases are missing from the full $174 \times 4 = 696$ configurations due to processing errors, and the simulations were not repeated.

\subsubsection{Drag-Outlier Split}
This is an alternative split of the same 174 geometries from the stratified split, in which 50\% of the validation and test sets are filled with the highest- and lowest-drag geometries, by round-robin assignment of the runs with the highest $C_d$ deviation from the mean. The rest of these sets are then composed of randomly assigned geometries, to obtain a final 80/10/10 split. This distribution probes performance at the edges of the aerodynamic envelope. Its training set is a different subset of the geometries, so every model is retrained from scratch on it under the same configuration as for the stratified split. Its benchmark results are reported in Section~\ref{subsec:ood}.

\subsubsection{Interpolated Split}
This split replaces the test set of the stratified split with the 20 geometries built from the 15 additional profiles of the interpolated set (Section~\ref{sec:sampling}), giving 80 test simulations on profile shapes that appear in no training geometry, while leaving the training and validation sets untouched. Because its training and validation sets are identical to those of the stratified split, no retraining is required or performed. These geometries are evaluated with the checkpoints trained on the stratified split. This split therefore probes generalization to unseen profile shapes. Its benchmark results are also reported in Section~\ref{subsec:ood}.

Table~\ref{tab:split_sizes} lists the contents of each split, in terms of number of total simulations and of unique geometries.

\begin{table}[hbt!]
\caption{\label{tab:split_sizes} The number of simulations in each of the three splits, with the number of unique geometries in parentheses.}
\centering
\begin{tabular}{lcccl}
\toprule
Split & Training & Validation & Test & Purpose \\
\midrule
Stratified    & 547 (138) & 72 (18) & 71 (18) & Training and primary benchmark \\
Drag-outlier  & 558 (141) & 68 (17) & 64 (16) & Extremes of the aerodynamic envelope \\
Interpolated  & 547 (138) & 72 (18) & 80 (20) & Unseen airfoil profiles \\
\bottomrule
\end{tabular}
\end{table}

\subsection{Surrogate Models}
\label{subsec:models}

The six architectures below span the three paradigms currently used for large-scale surface-field prediction: convolutional neural operators, graph neural networks, and transformers. For each model, the description states the paradigm it belongs to, the mechanism by which it encodes geometry and produces localized predictions, and, where one was needed, the adaptation made to apply it to this dataset.

\begin{enumerate}
    \item \textbf{DoMINO} (NVIDIA) \cite{domino}: A convolutional neural operator that processes 3D point clouds using spherical point convolution kernels at multiple scales. Geometric features are encoded into a structured latent representation, from which localized predictions are made via stencil-based aggregation. For this dataset, explicit inclusion of the wind-tunnel moving belt in the training data was required to correctly encode ride-height variations, as the model otherwise struggled to distinguish different ride heights of the same wing geometry.
    \item \textbf{X-MeshGraphNet} (NVIDIA) \cite{xmeshgraphnet}, an extension of MeshGraphNet \cite{meshgraphnet}: A graph neural network (GNN) that constructs a custom graph directly from the wing CAD surface, with message-passing layers propagating information between nodes across multiple spatial scales. Custom preprocessing was developed for this dataset, including cell-normal-based edge filtering, to prevent nonphysical information leakage across thin wing sections and sharp corners.
    \item \textbf{AB-UPT} (JKU Linz) \cite{abupt}: A transformer-based neural operator using an anchor-branched attention mechanism that reduces the computational complexity of attention from $O(N^2)$ to $O(N)$ by computing keys and values only from a small subset of anchor points. Separate geometry, surface, and volume branches communicate via cross-attention. For the benchmark it is run in a surface-only configuration, with the volume branch disabled. The variant trained with the addition of volumetric targets is reported in Appendix~\ref{sec:appendix_abupt_volume}.
    \item \textbf{AeroJEPA} \cite{aerojepa}: A joint-embedding predictive architecture (JEPA) model that learns semantic latent representations for aerodynamic field modeling. It encodes geometry with a context encoder and a conditioned predictor, then calculates the latent loss against the outputs of the target encoder, which encodes the ground-truth flow variables. A separate implicit neural representation (INR) decoder provides flow-field predictions at arbitrary resolutions.
    \item \textbf{GeoTransolver} (NVIDIA) \cite{geotransolver}: A geometry-aware evolution of Transolver \cite{transolver} that adds geometry-aware latent embedding (GALE) attention, in the spirit of DoMINO, to inject local and global geometric context. In addition, it persistently reprojects geometry and boundary-condition information into the physical state space at every transformer block via cross-attention.
    \item \textbf{GAOT} (ETH Zurich) \cite{gaot}: A geometry-aware operator transformer that uses a multiscale attentional graph neural operator (MAGNO) as encoder and decoder, together with geometry embeddings derived from local statistical shape descriptors. A vision-transformer (ViT) processor operates on an equispaced latent token grid for computational efficiency, while the MAGNO decoder still enables evaluation at arbitrary query points.
\end{enumerate}

All six models are trained to predict the surface fields $C_p$ and $\bm{w_s}$ from the wing geometry, which is the task the benchmark compares them on, and every one of them is trained on those surface fields alone. Where an architecture supports combined surface and volumetric training, it was run in its surface-only configuration. AB-UPT was customized to remove the volumetric branch, as its default configuration is designed to use it. To verify reproducibility, the original model was also trained on the splits described in this paper with otherwise identical settings. It is reported in Appendix~\ref{sec:appendix_abupt_volume} as additional information.

Integral aerodynamic coefficients $C_l$ and $C_d$ are obtained by numerically integrating the predicted surface fields over the wing and endplate surfaces.

The architecture-specific model settings were determined by a combination of hyperparameter optimization with Optuna \cite{optuna} and empirical testing on the validation set of the stratified split. The number of epochs and the learning-rate settings, in contrast, were deliberately left at the values found in each model's reference implementation, so that no architecture is advantaged by a better-tuned optimization schedule than its authors intended. Table~\ref{tab:model_config} reports the primary parameters of interest for the trained models, while Appendix~\ref{sec:appendix_models} reports the full configuration of each.

\begin{table}[hbt!]
\caption{\label{tab:model_config} Principal training configuration per model. ``Width'' is the latent or hidden dimension of the main processor, ``layers'' the depth of the processor, and ``heads'' the number of attention heads where attention is used. The point counts are per forward pass.}
\centering
\small
\begin{tabularx}{\textwidth}{lcccXcc}
\toprule
Model & Width & Layers & Heads & Sampled points & Epochs & Learning rate \\
\midrule
AB-UPT & 192 & 6 & 3 & \num{16384} surface, \num{2000} geometry & 500 & \num{5e-5} \\
GeoTransolver & 256 & 13 & 4 & \num{100000} surface, \num{50000} geometry & 500 & \num{3e-4} \\
AeroJEPA & 128 & 6 & 4 & \num{8192} surface, 512 tokens & 300 & \num{1e-3} \\
GAOT & 256 & 10 & 8 & $64\times32\times32$ latent grid & 300 & \num{8e-4} \\
DoMINO & 256 & UNet & --- & \num{16384} surface, \num{300000} geometry & 500 & \num{5e-4} \\
X-MeshGraphNet & 256 & 15 & --- & \num{40000} graph nodes & 500 & \num{1e-3} \\
\bottomrule
\end{tabularx}
\end{table}

\subsection{Error Metrics}
\label{subsec:metrics}

Throughout this section, a hat (\textasciicircum) denotes a predicted quantity and an unhatted symbol the corresponding CFD ground truth. Accuracy is reported at two levels: on the integral aerodynamic coefficients, which are what a design workflow ultimately consumes, and on the surface fields themselves, which are what the models actually regress.

For the coefficients, the primary metric is the absolute percentage error of a single simulation,
\begin{equation}
    \label{eq:coeff_err}
    \mathrm{Err}(C_i) = \left\lvert\frac{\hat{C}_{i}-C_i}{C_i}\right\rvert \times 100\%, \qquad i \in \{l, d\},
\end{equation}
of which the mean and the standard deviation over a test set are reported. Two coefficients of determination are reported alongside it, to evaluate whether models respect the order of the design space. The first,
\begin{equation}
    \label{eq:r2_fit}
    R^2_\mathrm{fit} = \operatorname{corr}\left(C_i, \hat{C}_i\right)^2,
\end{equation}
is the squared Pearson correlation between prediction and ground truth, equivalently the coefficient of determination of the least-squares line fitted through the predicted-versus-true scatter. It measures how well a model tracks the coefficient trend across the design space and is by construction insensitive to a constant offset or a systematic scale error. The second,
\begin{equation}
    \label{eq:r2_agreement}
    R^2_{1:1} = 1 - \frac{\sum_k \left(\hat{C}_{i,k}-C_{i,k}\right)^2}{\sum_k \left(C_{i,k}-\overline{C}_i\right)^2},
\end{equation}
with $k$ running over the simulations of the test set and $\overline{C}_i$ denoting their mean, is taken about the $1{:}1$ line rather than a fitted one and therefore penalizes bias and scale errors as well as scatter. A high $R^2_\mathrm{fit}$ combined with a lower $R^2_{1:1}$ highlights a prediction that is well correlated with the truth but offset from it.

For a field $\phi \in \{C_p, \lvert\bm{w_s}\rvert\}$ with element-wise error $e_j = \hat{\phi}_j - \phi_j$ over the $N$ surface elements of one simulation,
\begin{align}
    \label{eq:field_err}
    \mathrm{MSE}(\phi) = \frac{1}{N}\sum_{j=1}^{N} e_j^2,
    &&
    L_1(\phi) = \frac{\sum_{j=1}^{N} \lvert e_j\rvert}{\sum_{j=1}^{N} \lvert \phi_j \rvert},
    &&
    L_2(\phi) = \sqrt{\frac{\sum_{j=1}^{N} e_j^2}{\sum_{j=1}^{N} \phi_j^2}}.
\end{align}
The mean squared error (MSE) is left in the physical units of the field, while the two norm-based measures are normalized by the corresponding norm of the ground truth, which makes them dimensionless and therefore comparable between the dimensionless $C_p$ and the wall shear stress in \si{\pascal}. All three are computed per simulation and then averaged over the test set. The wall shear stress is evaluated on its magnitude. Every field metric reported below was recomputed element-wise from the prediction meshes themselves, with one implementation applied to all models.

\subsection{Benchmarking Results}
\label{subsec:results}

Table~\ref{tab:results} summarizes the coefficient accuracy of all six models on the test set of the stratified split (Section~\ref{subsec:splits}), ordered by mean $C_l$ error. Transformer-based architectures lead the benchmark, and the ranking is the same for both coefficients except for DoMINO, which moves up one place on $C_d$. Drag is the harder quantity for five of the six models, with the widest gaps for X-MeshGraphNet and AeroJEPA, whose $C_d$ error is $\simeq1.6$ times their $C_l$ error. The two coefficients of determination distinguish scatter from bias. GAOT reaches a high $R^2_\mathrm{fit}$ on $C_l$ ($0.989$) despite a higher mean error, so its predictions follow the correct trend and the error is concentrated in a small number of badly mispredicted cases. X-MeshGraphNet is the only model whose two $C_l$ scores differ appreciably ($0.961$ against $0.898$), the signature of a prediction that correlates well with the truth but is systematically offset from the $1{:}1$ line.

\begin{table}[hbt!]
\caption{\label{tab:results} Coefficient accuracy on the test simulations of the stratified split, ordered by mean $C_l$ error.}
\centering
\begin{tabular}{llcccccc}
\toprule
& & \multicolumn{3}{c}{$C_l$} & \multicolumn{3}{c}{$C_d$} \\
\cmidrule(lr){3-5}\cmidrule(lr){6-8}
Model & Architecture & Err. ($\mu, \sigma$), \% & $R^2_\mathrm{fit}$ & $R^2_{1:1}$ & Err. ($\mu, \sigma$), \% & $R^2_\mathrm{fit}$ & $R^2_{1:1}$ \\
\midrule
AB-UPT & Transformer & 0.85, 0.89 & 0.999 & 0.999 & 0.89, 0.74 & 0.999 & 0.999 \\
GeoTransolver & Transformer & 2.33, 2.81 & 0.995 & 0.994 & 3.15, 2.49 & 0.990 & 0.990 \\
AeroJEPA & JEPA & 2.83, 2.46 & 0.995 & 0.994 & 4.38, 3.01 & 0.987 & 0.977 \\
GAOT & Transformer & 5.62, 11.01 & 0.989 & 0.987 & 7.48, 6.51 & 0.963 & 0.957 \\
DoMINO & CNN & 7.05, 6.01 & 0.972 & 0.966 & 6.24, 3.93 & 0.977 & 0.971 \\
X-MeshGraphNet & GNN & 8.87, 5.53 & 0.961 & 0.898 & 13.80, 10.86 & 0.904 & 0.897 \\
\bottomrule
\end{tabular}
\end{table}

\begin{figure}[hbt!]
    \centering
    \includesvg[width=\linewidth]{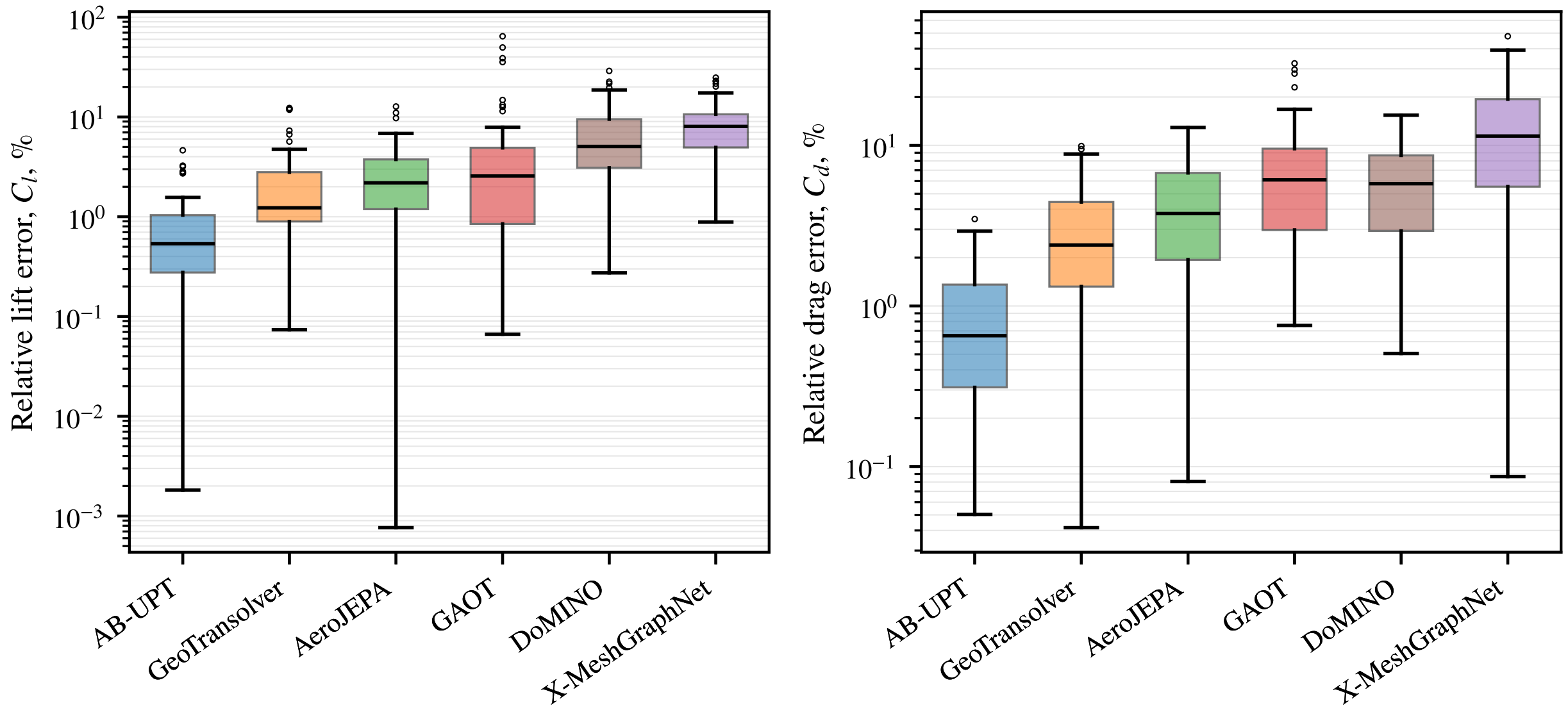}
    \caption{Distribution of per-simulation $C_l$ (left) and $C_d$ (right) relative error on the test set of the stratified split.}
    \label{fig:boxplot}
\end{figure}

\begin{figure}[hbt!]
    \centering
    \includesvg[width=\linewidth]{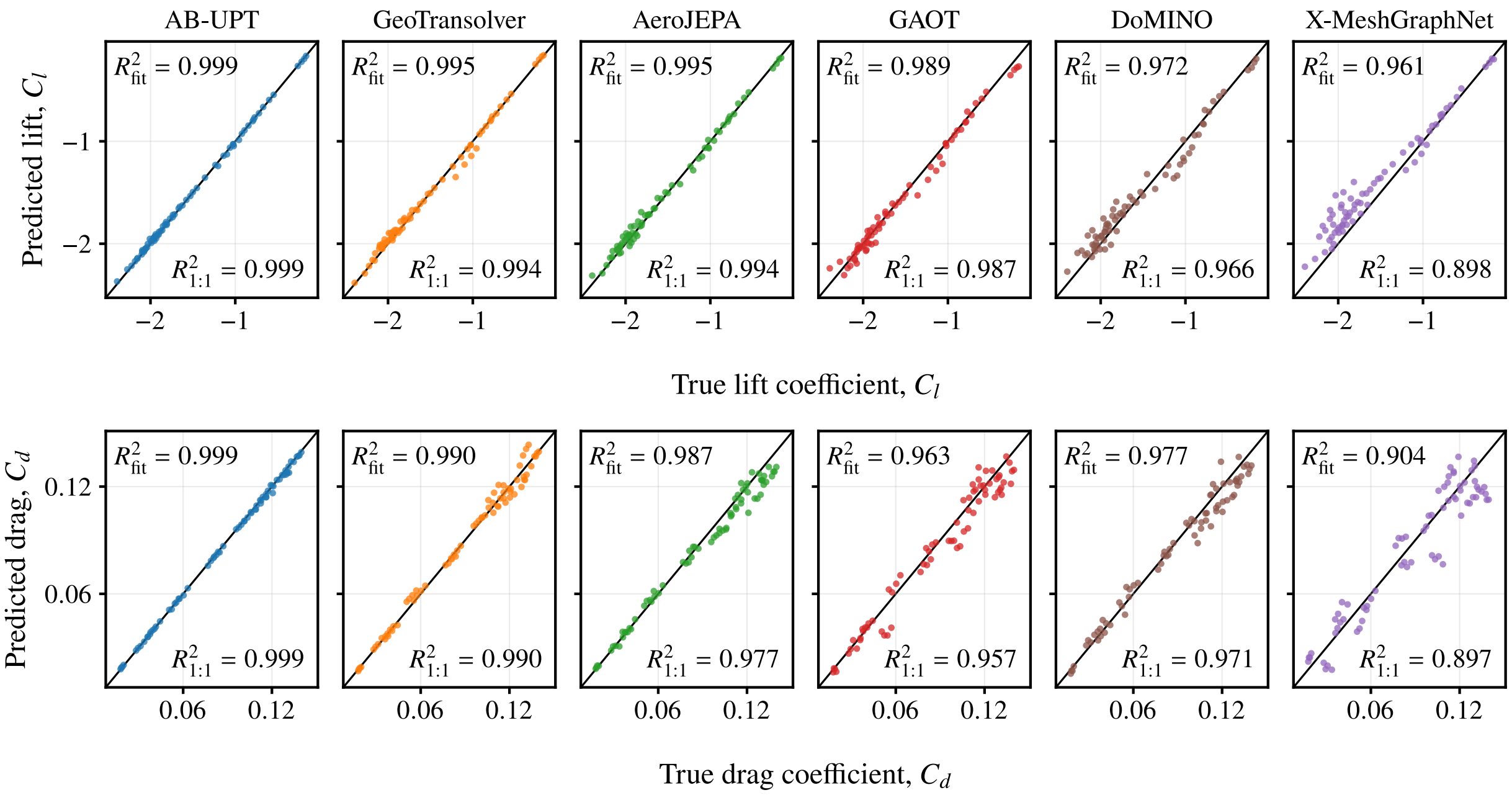}
    \caption{Predicted versus true $C_l$ (top row) and $C_d$ (bottom row) on the test set of the stratified split, with both coefficients of determination annotated in each panel. The solid diagonal is the $1{:}1$ line.}
    \label{fig:parity}
\end{figure}

Figure~\ref{fig:boxplot} shows the full per-simulation error distribution as box plots, confirming that AB-UPT's advantage holds not only in the mean but throughout the distribution, while GAOT and X-MeshGraphNet show markedly heavier tails, GAOT for $C_l$ and X-MeshGraphNet for $C_d$. Figure~\ref{fig:parity} plots the same predictions against the ground truth. It makes the two failure modes visible directly: GAOT's $C_l$ panel is tight around the $1{:}1$ line except for a handful of far outliers, whereas X-MeshGraphNet's scatter is broad and tilted, with the highest-downforce cases systematically underpredicted in magnitude and the highest-drag cases overpredicted.

\begin{figure}[hbt!]
    \centering
    \includesvg[width=\linewidth]{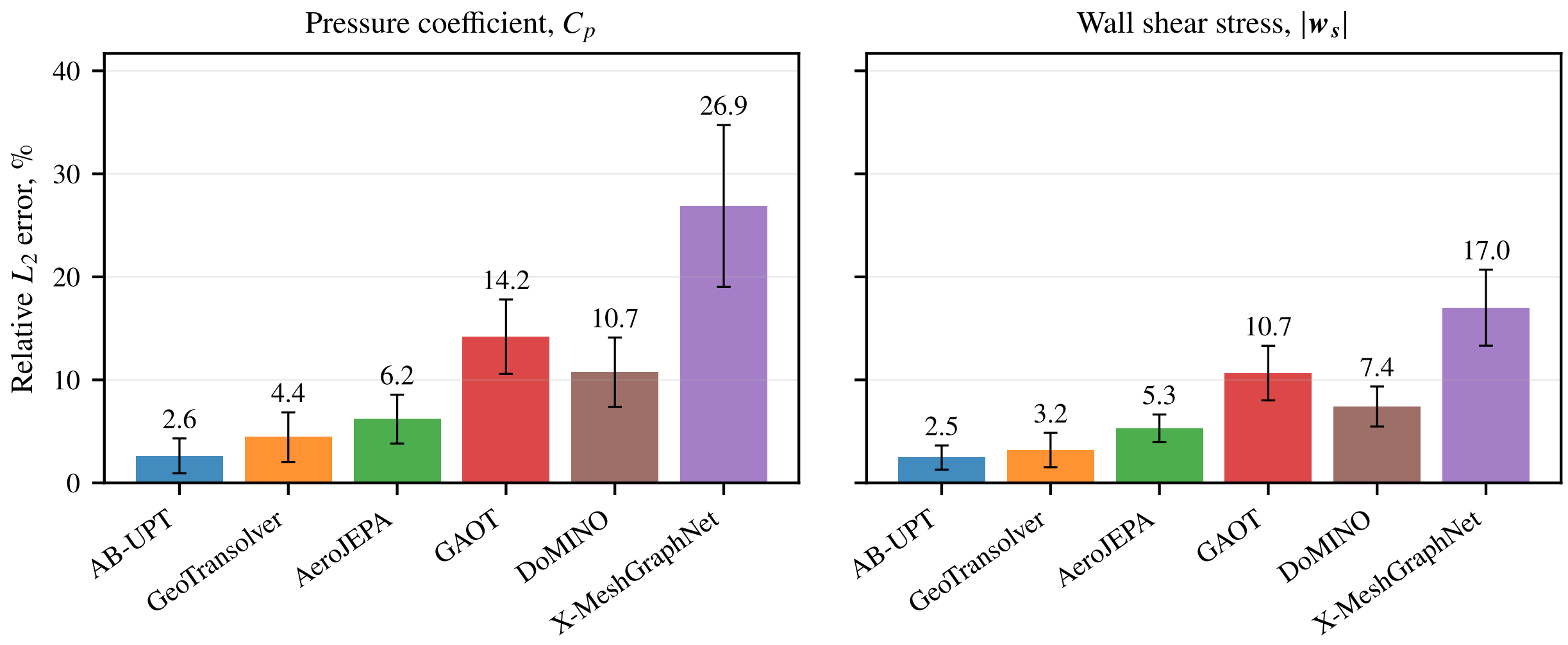}
    \caption{Mean relative $L_2$ error of the predicted surface fields on the test set of the stratified split. Error bars span one standard deviation across the test simulations.}
    \label{fig:field_error}
\end{figure}

Integral coefficients are the quantities a design workflow ultimately consumes, but they are integrals: cancellation between over- and underpredicted regions can hide substantial local error. Table~\ref{tab:field_errors} therefore reports the surface-field metrics of Eq.~\eqref{eq:field_err} on the same test set, and Fig.~\ref{fig:field_error} shows the relative $L_2$ error with its spread across the test set. The field ranking reproduces the coefficient ranking model for model, with one exception: DoMINO, fifth on $C_l$, is fourth on both fields, ahead of GAOT, which is consistent with its comparatively strong $C_d$ prediction. The agreement between the two levels indicates that the coefficient errors of Table~\ref{tab:results} are not an artifact of favorable cancellation. For every model, the relative $L_2$ error exceeds the relative $L_1$ error, by between $10\%$ and $32\%$ on $C_p$ and by between $26\%$ and $62\%$ on the wall shear stress. This places the field error predominantly in localized regions, such as the suction peak, the endplate junctions, and the wake-facing side of the profile, rather than spreading it evenly over the surface.

\begin{table}[hbt!]
\caption{\label{tab:field_errors} Surface-field accuracy on the test set of the stratified split. All quantities are averaged over the test set.}
\centering
\begin{tabular}{lcccccc}
\toprule
& \multicolumn{3}{c}{$C_p$} & \multicolumn{3}{c}{$\lvert\bm{w_s}\rvert$} \\
\cmidrule(lr){2-4}\cmidrule(lr){5-7}
Model & MSE & $L_1$, \% & $L_2$, \% & MSE, \si{\pascal\squared} & $L_1$, \% & $L_2$, \% \\
\midrule
AB-UPT & 0.0017 & 2.00 & 2.63 & 0.102 & 1.51 & 2.46 \\
GeoTransolver & 0.0042 & 3.57 & 4.44 & 0.184 & 2.28 & 3.19 \\
AeroJEPA & 0.0071 & 5.51 & 6.19 & 0.397 & 3.95 & 5.30 \\
GAOT & 0.0285 & 12.56 & 14.19 & 1.533 & 7.97 & 10.66 \\
DoMINO & 0.0179 & 9.77 & 10.75 & 0.775 & 5.87 & 7.42 \\
X-MeshGraphNet & 0.1022 & 21.23 & 26.89 & 3.946 & 11.94 & 17.01 \\
\bottomrule
\end{tabular}
\end{table}

\subsection{Interpolated and Drag-Outlier Test Sets}
\label{subsec:ood}

\begin{figure}[hbt!]
    \centering
    \includesvg[width=\linewidth]{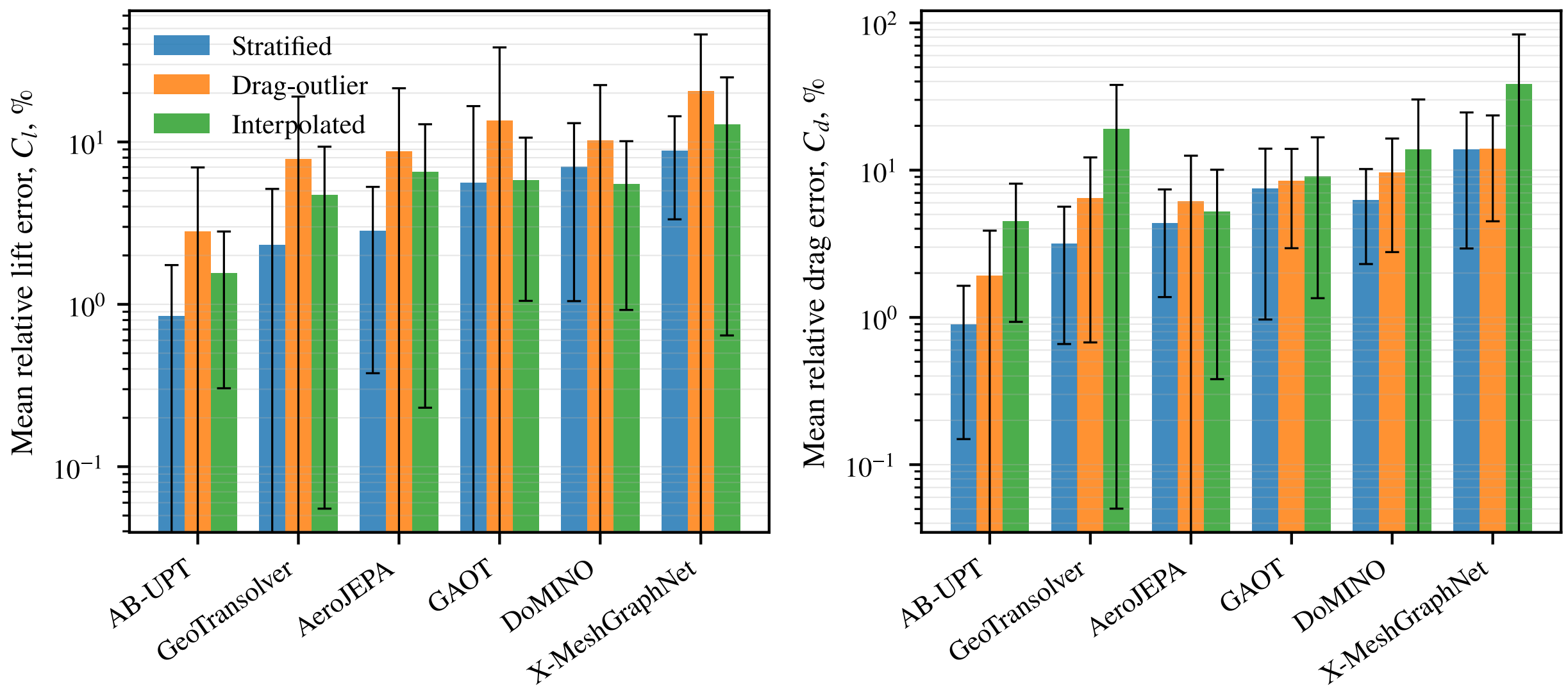}
    \caption{Mean relative $C_l$ (left) and $C_d$ (right) error of every model on the three test sets. Error bars span one standard deviation across the test simulations. Note the logarithmic ordinate.}
    \label{fig:split_comparison}
\end{figure}

Table~\ref{tab:results_splits} reports the same coefficient and field metrics on the two additional test sets of Section~\ref{subsec:splits}, and Fig.~\ref{fig:split_comparison} places all three test sets side by side. The two sets are evaluated differently, as described in Section~\ref{subsec:splits}. The interpolated set of profiles is predicted by the checkpoints trained on the stratified split, whereas the drag-outlier split has its own training set and every model was retrained on it. Both additional sets are harder than the stratified one for every model, but with two separate failure modes.

The drag-outlier set, whose validation and test geometries are drawn from the extremes of the drag distribution, hurts lift far more than drag: every model's $C_l$ error grows by a factor of between $1.5$ and $3.4$, while the $C_d$ error grows more slowly, by at most a factor of $2.2$. The only exception is DoMINO, where the trend reverses. For most models, the standard deviations grow at least as fast as the means, so the degradation is dominated by a minority of extreme geometries rather than by a uniform loss of accuracy. As expected, extrapolation outside of the training boundary produces worse results than mere interpolation.

The interpolated set, built from profile shapes that appear in no training geometry, reverses the pattern. Lift remains comparatively well predicted, whereas drag degrades sharply.

\begin{table}[hbt!]
\caption{\label{tab:results_splits} Performance on the drag-outlier and interpolated test sets.}
\centering
\small
\begin{tabular}{lcccccccc}
\toprule
& \multicolumn{4}{c}{Drag-outlier} & \multicolumn{4}{c}{Interpolated} \\
\cmidrule(lr){2-5}\cmidrule(lr){6-9}
& $C_l$ Err. & $C_d$ Err. & $C_p$ & $\lvert\bm{w_s}\rvert$ & $C_l$ Err. & $C_d$ Err. & $C_p$ & $\lvert\bm{w_s}\rvert$ \\
Model & ($\mu, \sigma$), \% & ($\mu, \sigma$), \% & $L_2$, \% & $L_2$, \% & ($\mu, \sigma$), \% & ($\mu, \sigma$), \% & $L_2$, \% & $L_2$, \% \\
\midrule
AB-UPT & 2.82, 4.15 & 1.92, 1.96 & 4.18 & 3.76 & 1.56, 1.25 & 4.51, 3.58 & 5.00 & 3.85 \\
GeoTransolver & 7.89, 11.14 & 6.44, 5.76 & 9.10 & 5.52 & 4.70, 4.65 & 18.98, 18.93 & 13.47 & 6.41 \\
AeroJEPA & 8.75, 12.67 & 6.14, 6.39 & 10.38 & 7.72 & 6.54, 6.31 & 5.22, 4.84 & 11.17 & 7.55 \\
GAOT & 13.58, 24.65 & 8.45, 5.50 & 16.38 & 11.32 & 5.84, 4.79 & 9.03, 7.68 & 17.35 & 11.95 \\
DoMINO & 10.25, 12.16 & 9.58, 6.81 & 14.57 & 9.39 & 5.52, 4.59 & 13.80, 16.40 & 15.02 & 8.24 \\
X-MeshGraphNet & 20.58, 25.34 & 14.01, 9.52 & 32.92 & 19.63 & 12.82, 12.17 & 38.37, 45.06 & 29.83 & 18.14 \\
\bottomrule
\end{tabular}
\end{table}

\subsection{Directional Consistency of Predicted Deltas}
\label{subsec:delta}

\begin{figure}[hbt!]
    \centering
    \includesvg[width=0.72\linewidth]{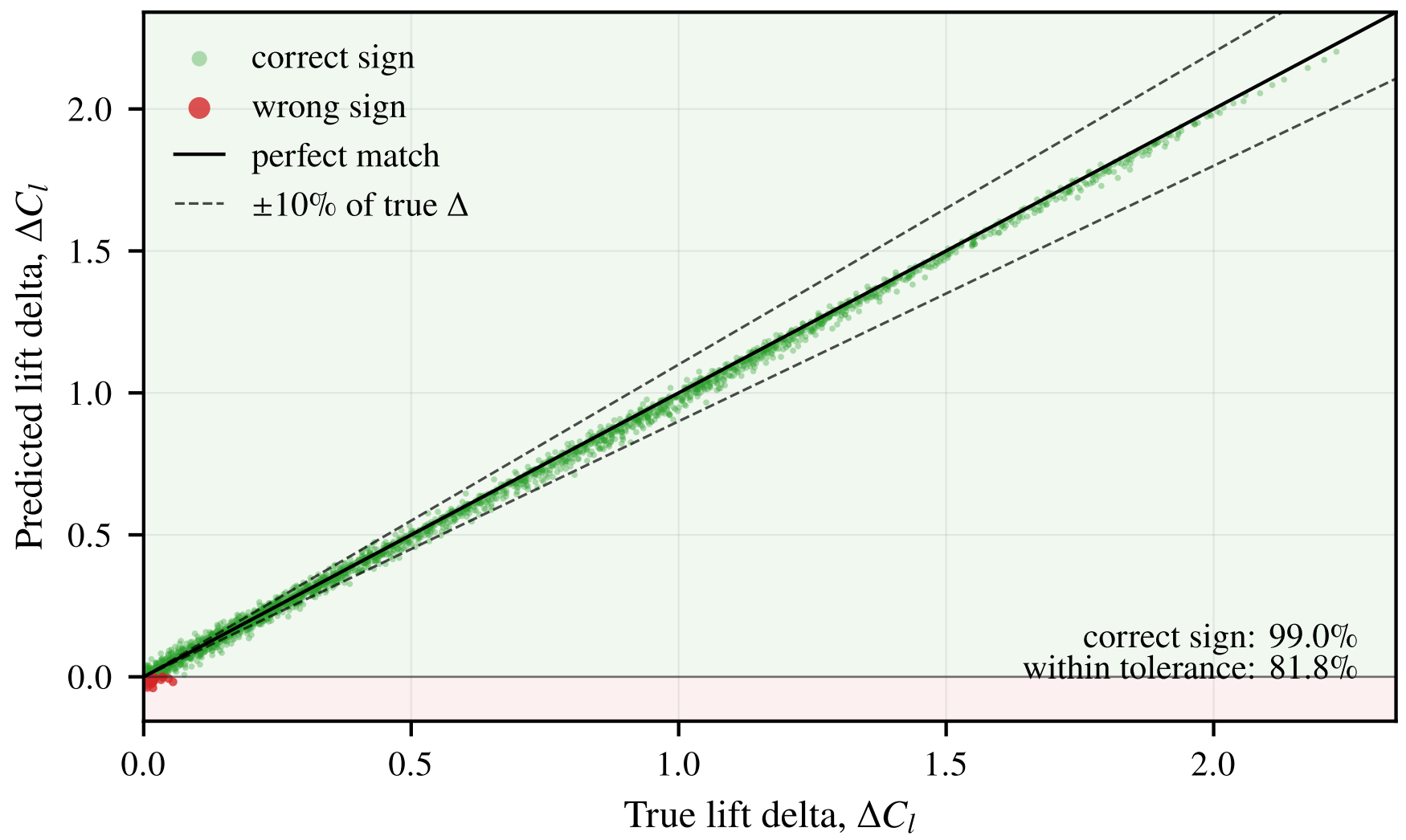}
    \caption{AB-UPT: predicted vs.\ true $C_l$ deltas between all pairs of test-set simulations.}
    \label{fig:delta_abupt}
\end{figure}

\begin{figure}[hbt!]
    \centering
    \includesvg[width=0.72\linewidth]{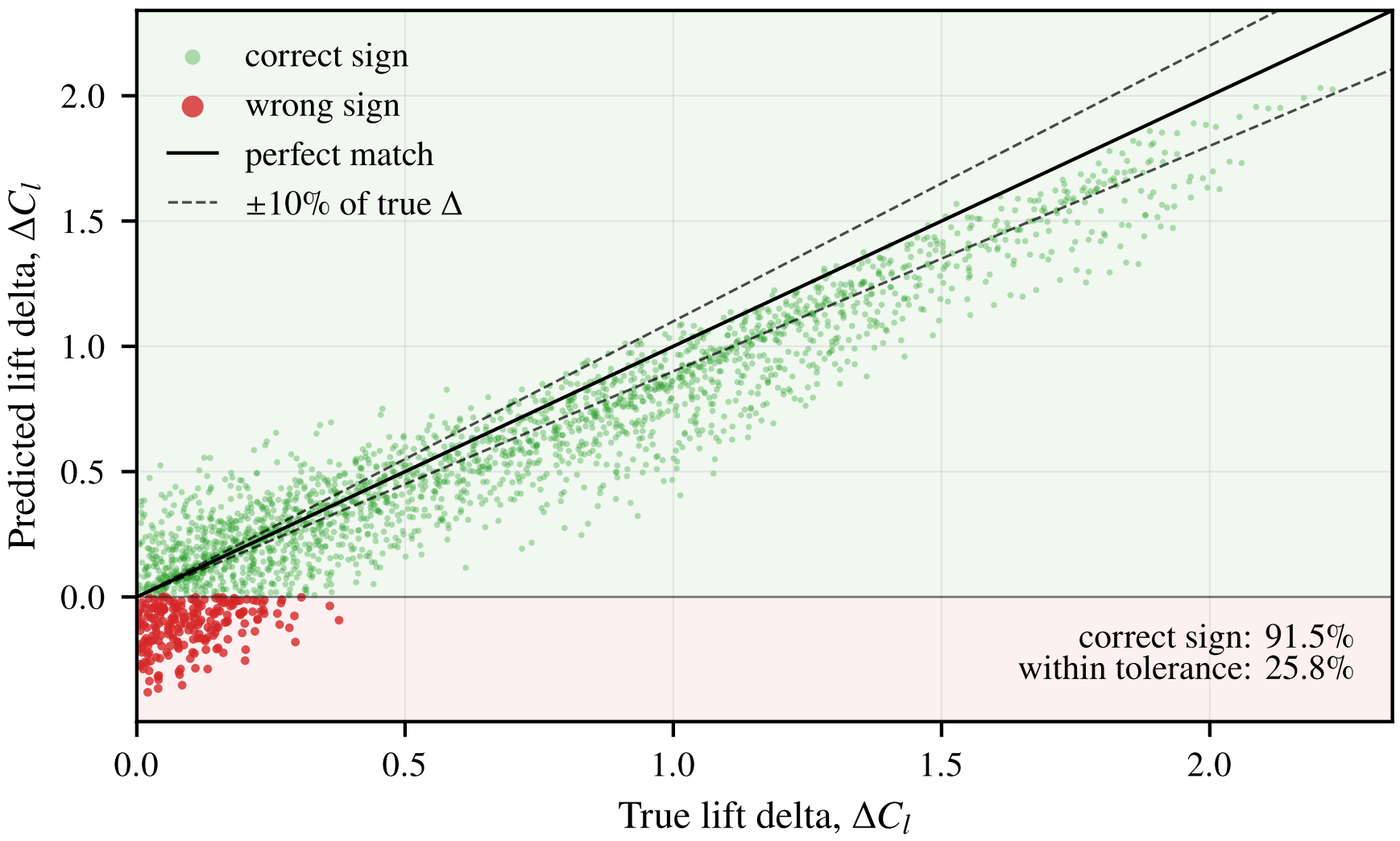}
    \caption{X-MeshGraphNet: predicted vs.\ true $C_l$ deltas between all pairs of test-set simulations.}
    \label{fig:delta_xmgn}
\end{figure}

Beyond the absolute prediction error, a practically relevant question for a surrogate model used in an iterative design workflow is whether it correctly predicts the direction of change in $C_l$ and $C_d$ between two related geometries, even if the absolute magnitude carries some error. For every ordered pair of the stratified test-set simulations, we compute the true and predicted coefficient deltas relative to one another and classify each pair as a directional match if $\mathrm{sign}(\Delta \hat{C}_i) = \mathrm{sign}(\Delta C_i)$. Figure~\ref{fig:delta_abupt} shows the resulting diagram for AB-UPT's $C_l$ predictions, and Fig.~\ref{fig:delta_xmgn} for X-MeshGraphNet, illustrating the gap between the best- and worst-performing models of Table~\ref{tab:results}. AB-UPT reproduces the correct sign of the delta in $99.0\%$ of pairs and keeps the predicted delta within $10\%$ of that pair's true delta in $81.8\%$ of them, while X-MeshGraphNet reproduces the correct sign in $91.5\%$ of pairs, meets the same relative tolerance in only $25.8\%$, and shows visibly larger scatter around the perfect-match line. On $C_l$, the other models lie between these two at $97.7\%$ (GeoTransolver), $96.5\%$ (AeroJEPA and GAOT, tied), and $93.9\%$ (DoMINO). Directional agreement on $C_d$ is lower than on $C_l$ for five of the six models and ranges from $98.9\%$ for AB-UPT down to $85.3\%$ for X-MeshGraphNet.

\subsection{Predicted Field Visualizations}
\label{subsec:fields}

A necessary addition to the metrics of the preceding sections is an evaluation of the spatial distribution of the errors on the wing surface. This subsection therefore shows the local surface pressure error on one reference case: test run \texttt{0169\_7575}, a geometry from the stratified test set at the \SI{75}{\milli\meter} ride height. It is chosen as a representative sample, as its median relative $C_l$ error across the six models, $2.8\%$, coincides with the median across the whole test set. Every panel is viewed from below, looking towards $+Z$, so the visible surface is the suction side that generates most of the downforce and has the highest $C_p$ magnitude.

\begin{figure}[hbt!]
    \centering
    \subfloat[CFD ground truth, $C_p$]{\includegraphics[width=0.48\linewidth]{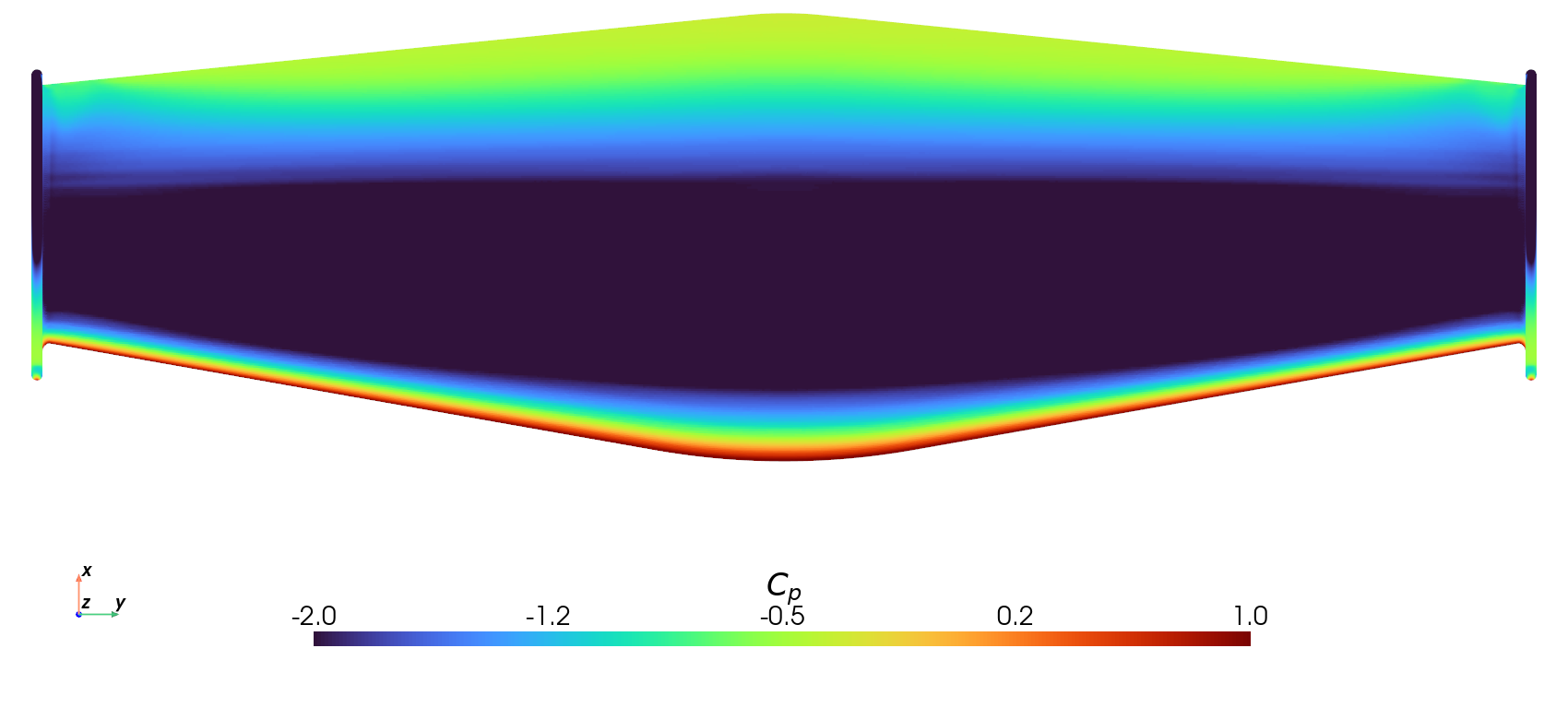}}\hfill
    \subfloat[AB-UPT prediction, $C_p$]{\includegraphics[width=0.48\linewidth]{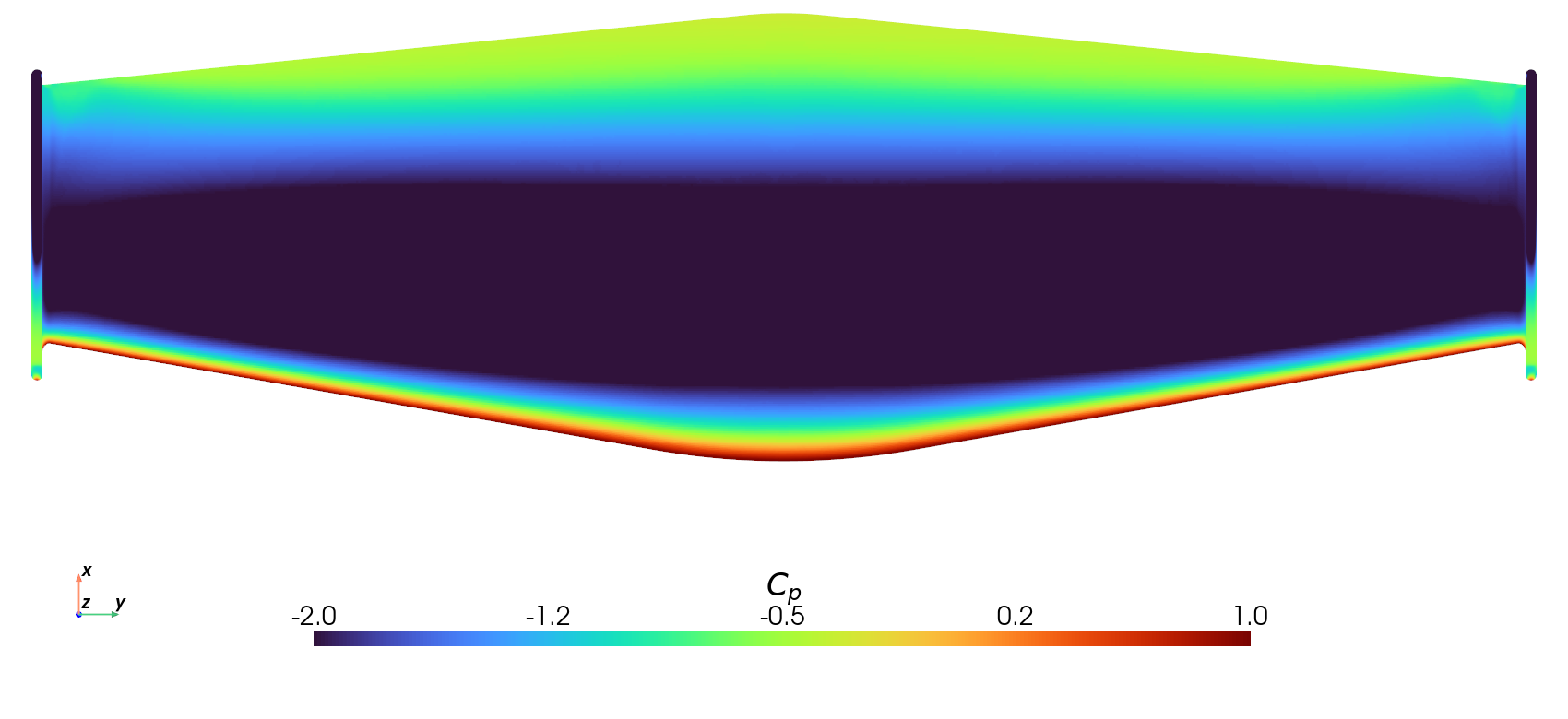}}\\
    \subfloat[{Signed error $\Delta C_p$, range $[-0.3,0.3]$}]{\includegraphics[width=0.9\linewidth]{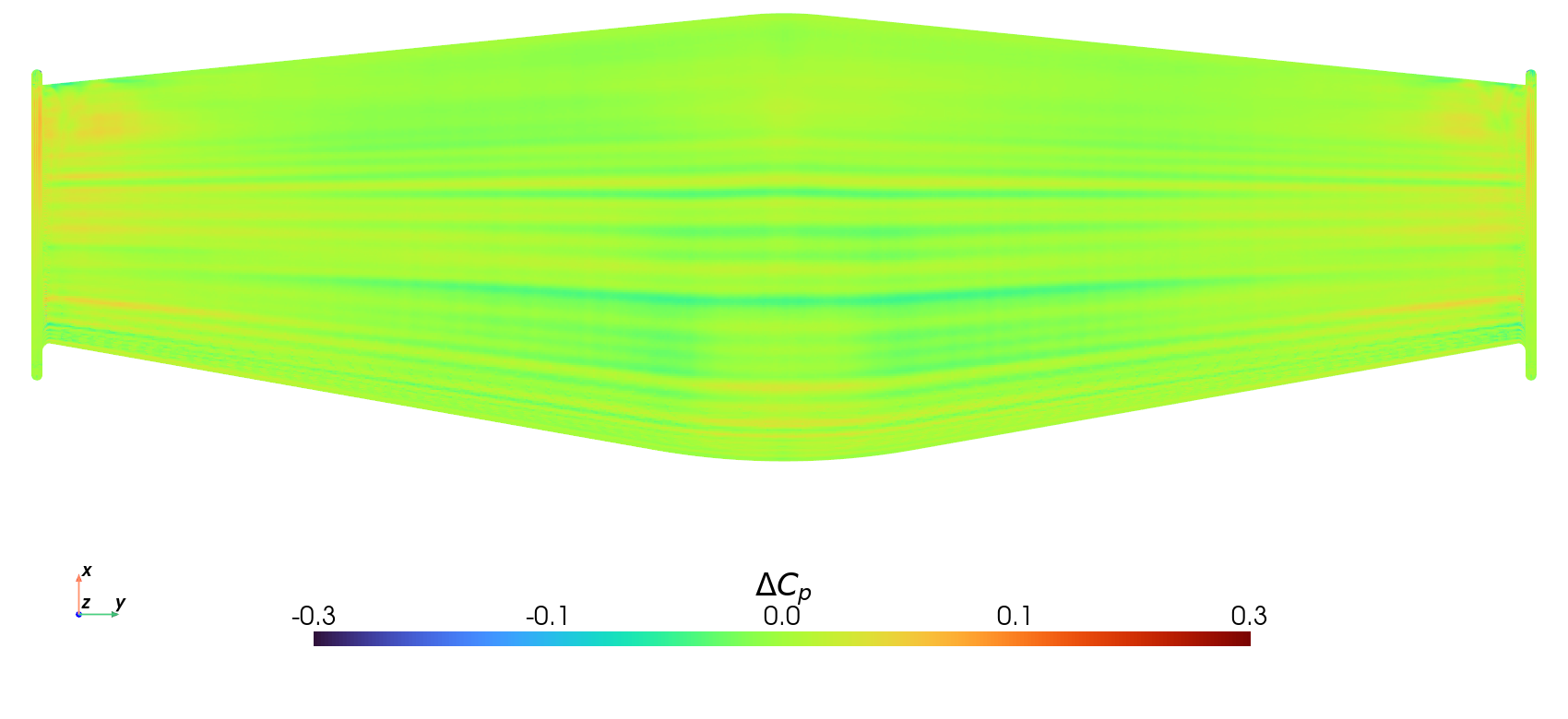}}
    \caption{Surface $C_p$ prediction and local error of AB-UPT on test run \texttt{0169\_7575} (stratified test set, \SI{75}{\milli\meter} ride height). Viewed from below, looking towards $+Z$.}
    \label{fig:screenshot_detail}
\end{figure}

\begin{figure}[hbt!]
    \centering
    \subfloat[AB-UPT]{\includegraphics[width=0.48\linewidth]{images/main/fields/cp_err_abupt.png}}\hfill
    \subfloat[GeoTransolver]{\includegraphics[width=0.48\linewidth]{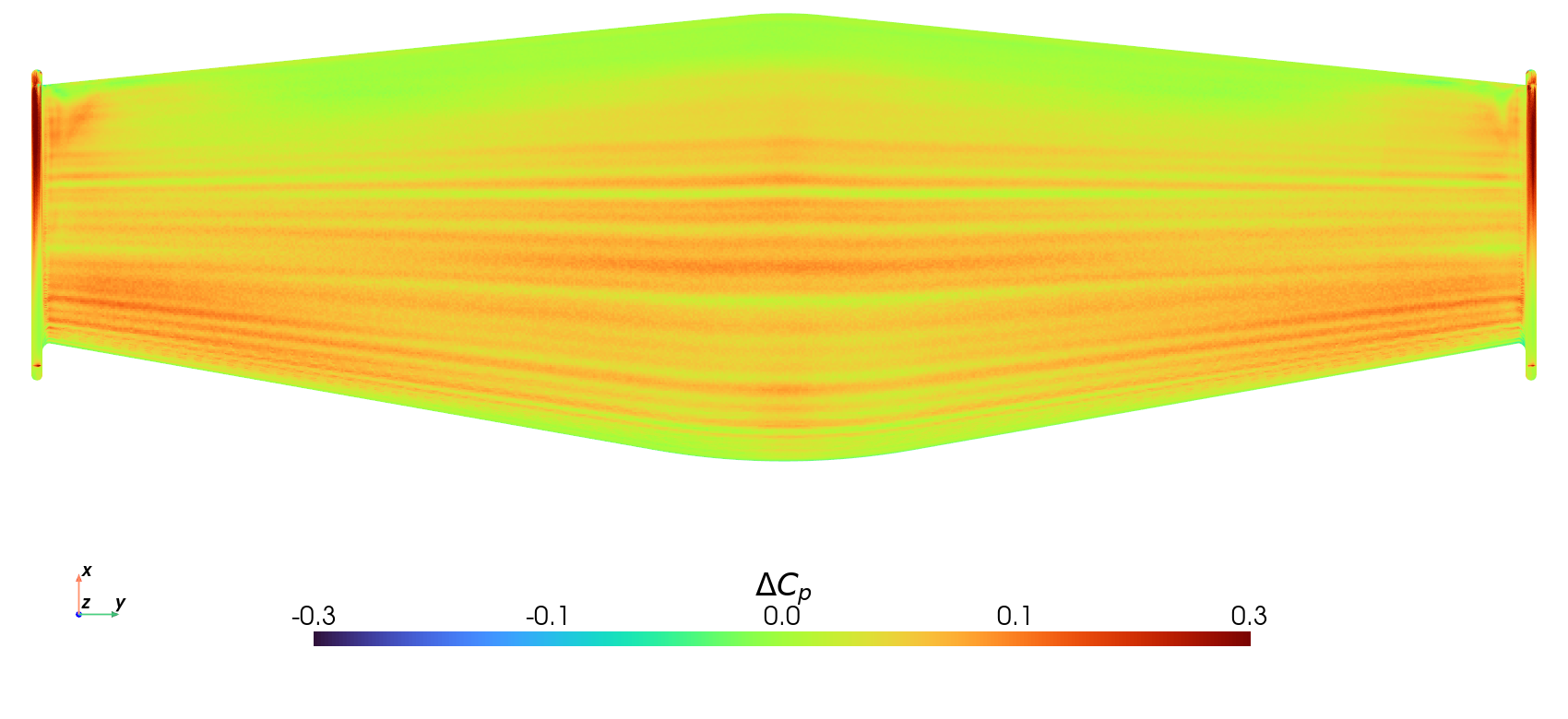}}\\
    \subfloat[AeroJEPA]{\includegraphics[width=0.48\linewidth]{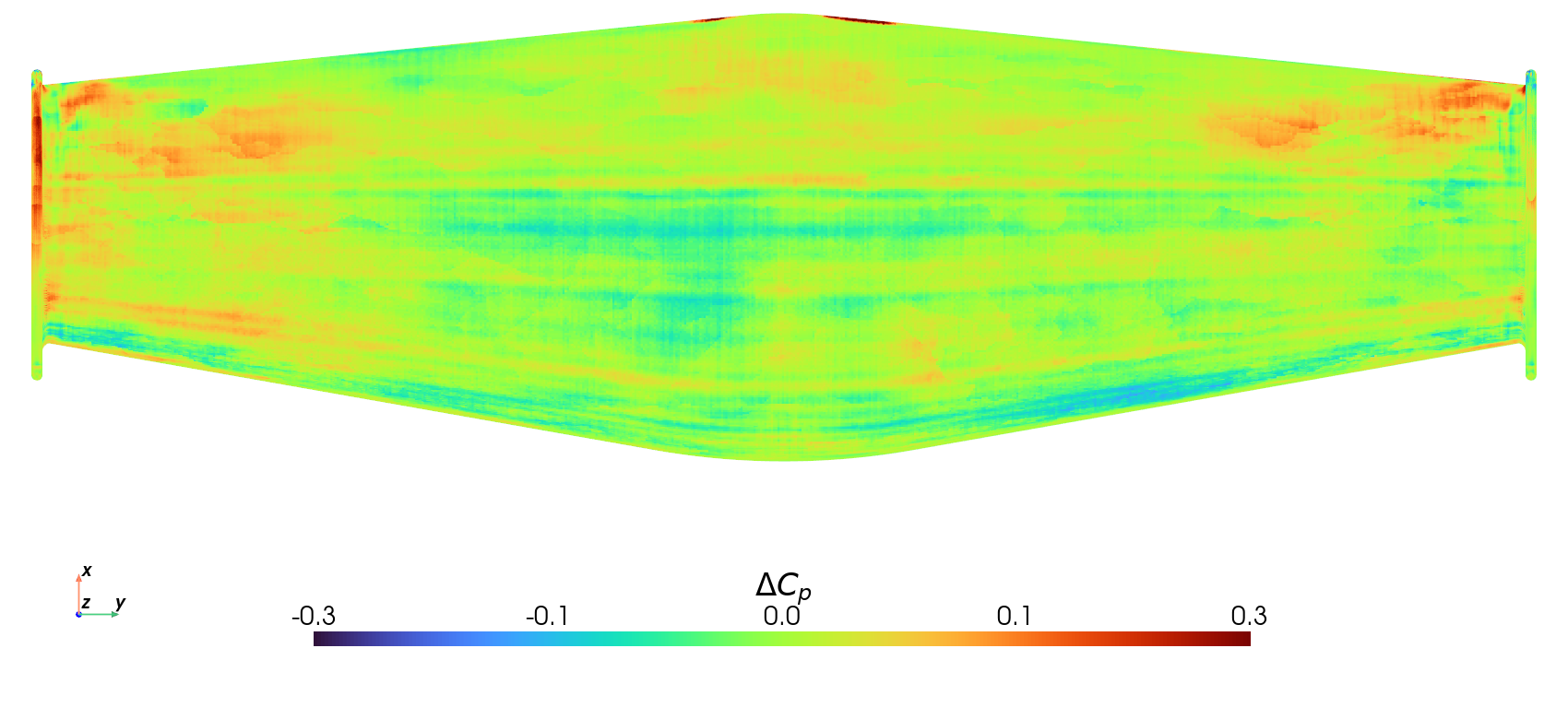}}\hfill
    \subfloat[GAOT]{\includegraphics[width=0.48\linewidth]{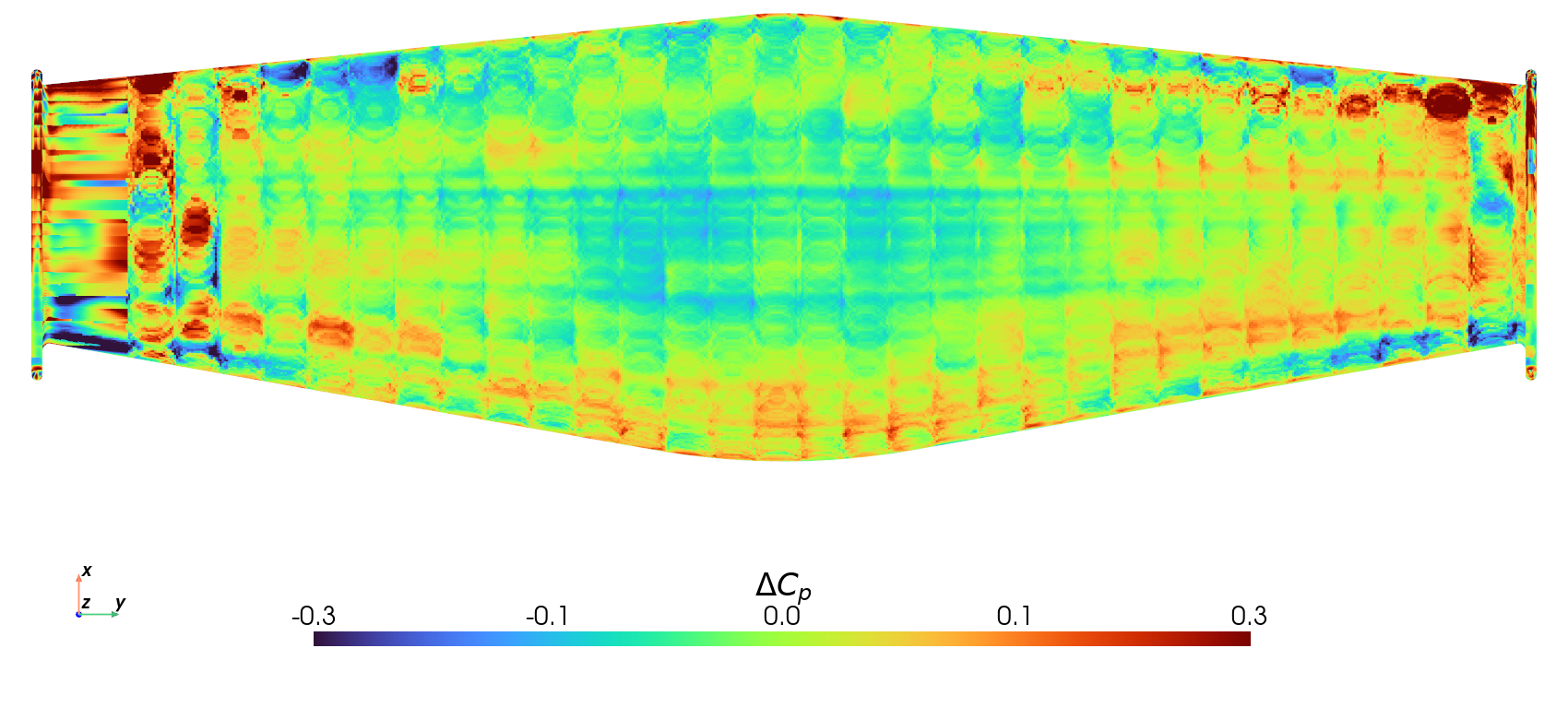}}\\
    \subfloat[DoMINO]{\includegraphics[width=0.48\linewidth]{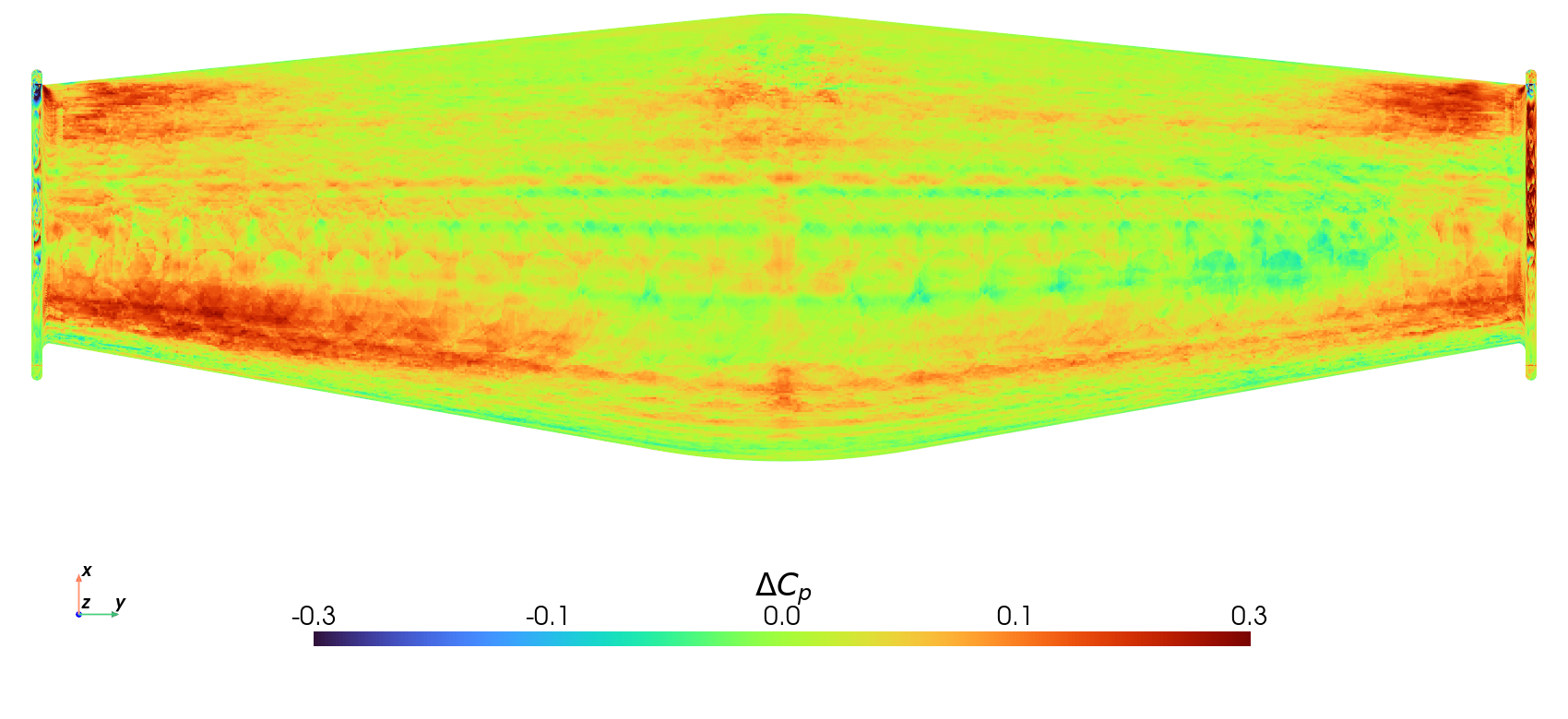}}\hfill
    \subfloat[X-MeshGraphNet]{\includegraphics[width=0.48\linewidth]{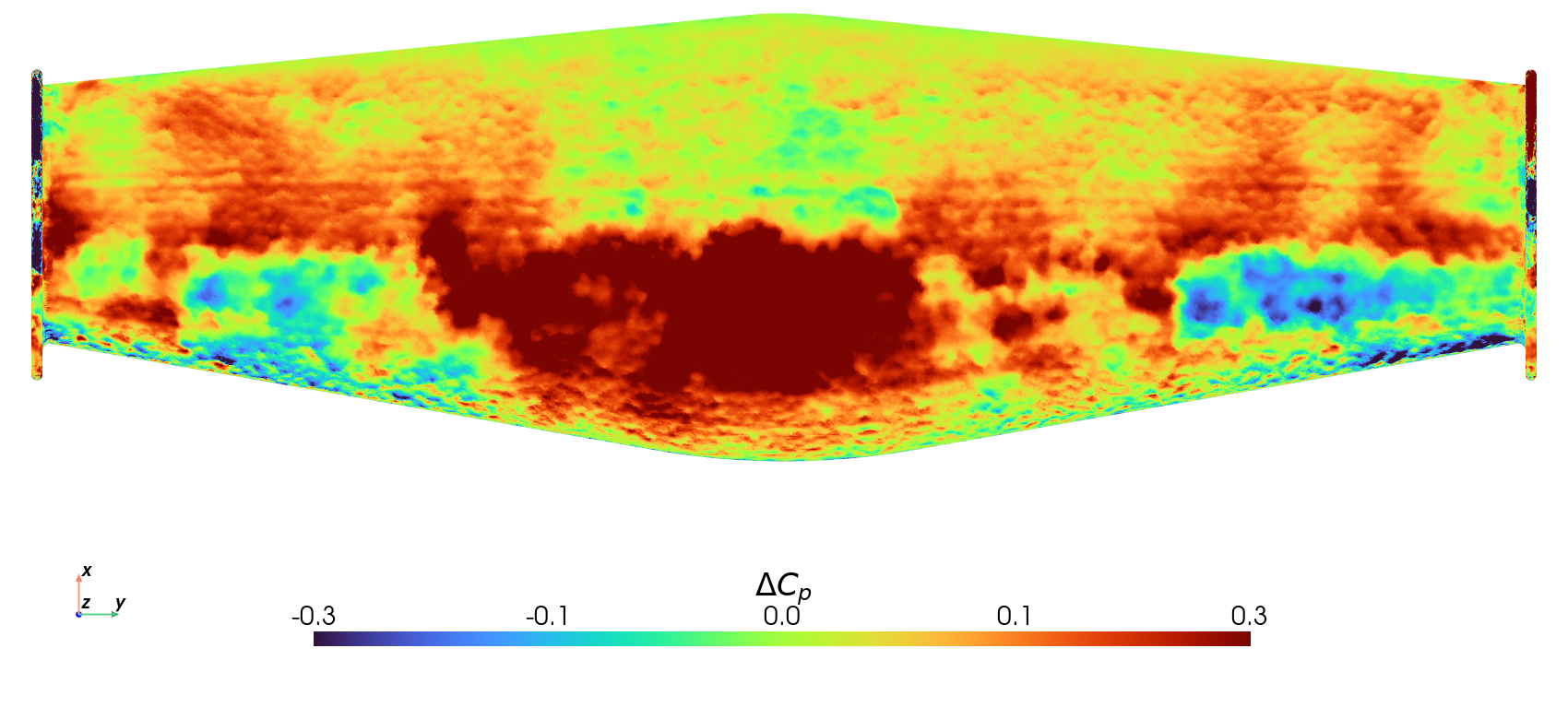}}
    \caption{Signed surface pressure error $\Delta C_p = C_p^\mathrm{pred}-C_p^\mathrm{CFD}$ on test run \texttt{0169\_7575}. Viewed from below, looking towards $+Z$.}
    \label{fig:screenshot_error}
\end{figure}

Figure~\ref{fig:screenshot_detail} places the $C_p$ prediction of the best-performing model, AB-UPT, next to the CFD ground truth together with its signed error, and Fig.~\ref{fig:screenshot_error} shows the signed error $\Delta C_p = C_p^\mathrm{pred}-C_p^\mathrm{CFD}$ of all six models.

In Fig.~\ref{fig:screenshot_detail}, the AB-UPT prediction reproduces the main features of the CFD ground truth: the spanwise suction band, the stagnation line along the leading edge, and the recovery towards the endplates. Together with the $R^2$ values of Table~\ref{tab:results}, this indicates that the models differ in the accuracy of the field rather than in its topology. The error maps of Fig.~\ref{fig:screenshot_error} separate them. AB-UPT is close to uniform, with only faint chordwise banding. GeoTransolver, AeroJEPA, GAOT, and DoMINO show progressively stronger error fields concentrated on the suction band, where the pressure gradients are largest. X-MeshGraphNet is qualitatively different, as its error distribution is affected by the graph construction on each geometry. Quantitatively, the error range of Fig.~\ref{fig:screenshot_error} contains all but $0.4\%$ of AB-UPT's surface elements on this run, against $14.5\%$ for X-MeshGraphNet.

\clearpage

\section{Reproducibility}
\label{sec:reproducibility}

Table~\ref{tab:checklist} maps the dataset and benchmark against the items of the ML Reproducibility Checklist v2.0.\footnote{Pineau, J., ``The Machine Learning Reproducibility Checklist,'' Version 2.0, 2020, \url{https://www.cs.mcgill.ca/~jpineau/ReproducibilityChecklist-v2.0.pdf}.}

\begin{table}[hbt!]
\caption{\label{tab:checklist} Dataset and benchmark status against the ML Reproducibility Checklist v2.0.}
\centering
\begin{tabularx}{\textwidth}{lX}
\toprule
Checklist item & Reference / Value \\
\midrule
Model/algorithm description & Section~\ref{subsec:models} \\
Model complexity analysis & Out of scope. Reported by the respective architecture papers \cite{domino,xmeshgraphnet,abupt,aerojepa,geotransolver,gaot} \\
Dataset statistics & Section~\ref{subsec:splits}, Table~\ref{tab:params} \\
Train/val/test split details & Section~\ref{subsec:splits}, Appendix~\ref{sec:appendix_splits}, Table~\ref{tab:split_sizes} \\
Preprocessing description & Section~\ref{sec:fileformat} \\
Downloadable dataset link & \url{https://huggingface.co/datasets/andreacenzato/Ground-Wing}\\
Dependency specification, training/evaluation code & Out of scope. Reported by the respective architecture papers \cite{domino,xmeshgraphnet,abupt,aerojepa,geotransolver,gaot} \\
Pretrained model weights & Out of scope. Can be regenerated by training the models with the given configurations. \\
Hyperparameters used & Appendix~\ref{sec:appendix_models}, Table~\ref{tab:model_config}. Selection method in Section~\ref{subsec:models} \\
Number of training/evaluation runs & Two per model (stratified and drag-outlier split) \\
Definition of reported metrics & Section~\ref{subsec:metrics}, Eqs.~\eqref{eq:coeff_err}--\eqref{eq:field_err} \\
Central tendency and variation of results & Tables~\ref{tab:results}, \ref{tab:field_errors}, and~\ref{tab:results_splits} \\
Excluded data & Below \\
Average runtime / energy cost & Out of scope. Training and inference cost per architecture is reported by the respective architecture papers \cite{domino,xmeshgraphnet,abupt,aerojepa,geotransolver,gaot} \\
Computing infrastructure description & NVIDIA H100 and A100 \\
\bottomrule
\end{tabularx}
\end{table}

For completeness, the following data are excluded from the reported results:

\begin{itemize}
    \item The airfoil profile used for the sensitivity analysis sweep is a proprietary internal design and cannot be shared; therefore, it does not appear in any simulation shared with this dataset.
    \item The mesh convergence, $Y^+$ study, and ride-height analysis were also performed using this proprietary profile.
\end{itemize}

\section{Conclusion}

This work introduces a parametric 3D wing-in-ground-effect CFD dataset, comprising 770 RANS simulations across 194 unique wing geometries and 4 ride heights, together with an associated benchmark of six neural surrogate architectures spanning convolutional, graph-based, and transformer paradigms. Transformer-based architectures dominate the benchmark, with AB-UPT achieving the lowest error on both aerodynamic coefficients ($0.85\%$ mean $C_l$ error, $0.89\%$ mean $C_d$ error), followed by GeoTransolver and AeroJEPA. The two additional test sets separate the failure modes cleanly, with the drag-outlier set degrading lift prediction and the interpolated profiles degrading drag prediction.

The dataset and its documentation, including the fully described CAD parametrization, sampling strategy, and CFD methodology (Section~\ref{sec:dataset}), are intended to serve as a standardized, ground-effect-specific benchmark for future data-driven aerodynamic prediction methods. The difficulty of the models in maintaining accuracy outside of the training envelope demonstrates that this dataset is a challenging benchmark evaluating model generalization.


\pagebreak

\appendix
\renewcommand{\thesubsection}{\thesection.\arabic{subsection}}
\titleformat{\subsection}
  {\normalsize\bfseries\singlespacing}
  {\thesection.\arabic{subsection}.\space}{0pt}{#1}[]

\section{Wing CAD Model and Geometry Constraints}
\label{sec:appendix_geometry}

In addition to the primary wing parameters (Table~\ref{tab:params}), the CAD model exposed the endplate parameters (Table~\ref{tab:endplate_params}) and the fillet parameters (Table~\ref{tab:fillet_params}). The first group is, by a wide margin, the most influential on the aerodynamic performance of the wing, which is why it is the only one varied across the dataset and the only one subject to the sensitivity analysis of Section~\ref{sec:sampling}. The other two groups are held constant at the values given below.

The endplate dimensions are derived from the ground-effect literature \cite{endplate1,endplate2}, but deviate from those references in two ways. First, the endplate extends rearward past the trailing edge of the profile to ensure CAD model stability. Second, the downward extension of the endplate is set by a fixed clearance of its lower edge from the ground (Table~\ref{tab:params}) rather than by a dimension relative to the wing. Keeping that lower edge at a constant height provides a clean definition for the ride height, at the cost of a small variation in total endplate area across geometries. The fillet radius joining the wing tip to the endplate was set to \SI{10}{\milli\meter} to prevent the triggering of local flow separation.

\begin{table}[hbt!]
\caption{\label{tab:endplate_params} Fixed endplate parameters.}
\centering
\begin{tabularx}{\textwidth}{lXc}
\toprule
Parameter & Description & Value \\
\midrule
Endplate UPWD & Vertical distance between the trailing edge at the tip and the top of the endplate & \SI{105}{\milli\meter} \\
Endplate RWD & Horizontal distance between the trailing edge at the tip and the rear edge of the endplate & \SI{20}{\milli\meter} \\
Endplate FWD & Horizontal distance between the leading edge at the tip and the front edge of the endplate & \SI{50}{\milli\meter} \\
Endplate thickness & Thickness of the endplate in the $y$ direction & \SI{15}{\milli\meter} \\
Endplate edge fillet & Radius of the fillet between the wing tip and the endplate & \SI{10}{\milli\meter} \\
\bottomrule
\end{tabularx}
\end{table}

The fillet parameters of Table~\ref{tab:fillet_params} exist purely to smooth the overall shape of the wing, in particular the transition between the two semispans at the centerline and the transition towards the wing tips. Their influence on the integral aerodynamic coefficients was found to be negligible compared to that of the primary parameters, so they were kept identical across all geometries.

\begin{table}[hbt!]
\caption{\label{tab:fillet_params} Fixed fillet parameters.}
\centering
\begin{tabularx}{\textwidth}{lXc}
\toprule
Parameter & Description & Value \\
\midrule
Leading-edge fillet & Radius of the fillet between the two semispans at the leading edge, in the $xy$ plane & \SI{1000}{\milli\meter} \\
Leading-edge center fillet & Radius of the fillet between the two semispans at the leading edge, in the $yz$ plane & \SI{500}{\milli\meter} \\
Leading-edge tip fillet & Radius of the fillet in the $yz$ plane between the leading edge at the tip and the $y$ direction & \SI{500}{\milli\meter} \\
Trailing-edge fillet & Radius of the fillet between the two semispans at the trailing edge, in the $xy$ plane & \SI{500}{\milli\meter} \\
Trailing-edge center fillet & Radius of the fillet between the two semispans at the trailing edge, in the $yz$ plane & \SI{500}{\milli\meter} \\
Trailing-edge tip fillet & Radius of the fillet in the $yz$ plane between the trailing edge at the tip and the $y$ direction & \SI{500}{\milli\meter} \\
\bottomrule
\end{tabularx}
\end{table}

Finally, the vertical positions at which the wing is simulated were set from the suspension ride heights of the reference vehicle, so that the setup could be defined with the same reference parameters used in full-vehicle work. Equations~\eqref{eq:rotation_center} and~\eqref{eq:rake_angle} below are written in terms of the suspension ride heights, as the rotation axis corresponds to the front axle of the reference vehicle. Configurations with a nonzero rake angle (i.e., the rotation of the chassis about the $y$ axis, positive when the rear of the vehicle sits higher than the front) were also simulated. Such a configuration is reproduced by rotating the wing about the point given by Eq.~\eqref{eq:rotation_center} by the angle of Eq.~\eqref{eq:rake_angle},
\begin{subequations}
    \begin{align}
        (x,y,z) &= (-\text{Front Axle Position},\ 0,\ \text{Front Ride Height}), \label{eq:rotation_center}\\
        \theta &= \arctan\left(\frac{\text{Rear Ride Height} - \text{Front Ride Height}}{\text{Wheelbase}}\right). \label{eq:rake_angle}
    \end{align}
\end{subequations}
Raked configurations were subsequently excluded from the final dataset. Since the wing is the only body present in the flow field, rotating it about the front axle is equivalent to changing its angle of attack, except for the rotation of the endplates. With the ride-height combinations of interest, Eq.~\eqref{eq:rake_angle} yields rake angles below \SI{1}{\degree}, making that difference negligible. Configurations with a nonzero roll angle were considered as well, but were deemed outside the scope of this work. The four flat ride heights of Table~\ref{tab:params} therefore span the relevant range of ground clearances without duplicating information already contained in the angle-of-attack parameter.

\section{Airfoil Profile Selection and Generation}
\label{sec:appendix_profiles}

The coordinates of every profile used in this work were obtained from the UIUC Airfoil Coordinates Database \cite{uiuc_airfoils}, the Airfoil Tools database \cite{airfoiltools}, or specific publications \cite{benzing,mshd}. The custom profile used on the front splitter of the reference race car was part of the original profile pool but is not part of this dataset, as it is proprietary. No coordinate file is available for it, which also precluded its use in the interpolation described below.

\begin{figure}[hbt!]
    \centering
    \includesvg[width=0.7\linewidth]{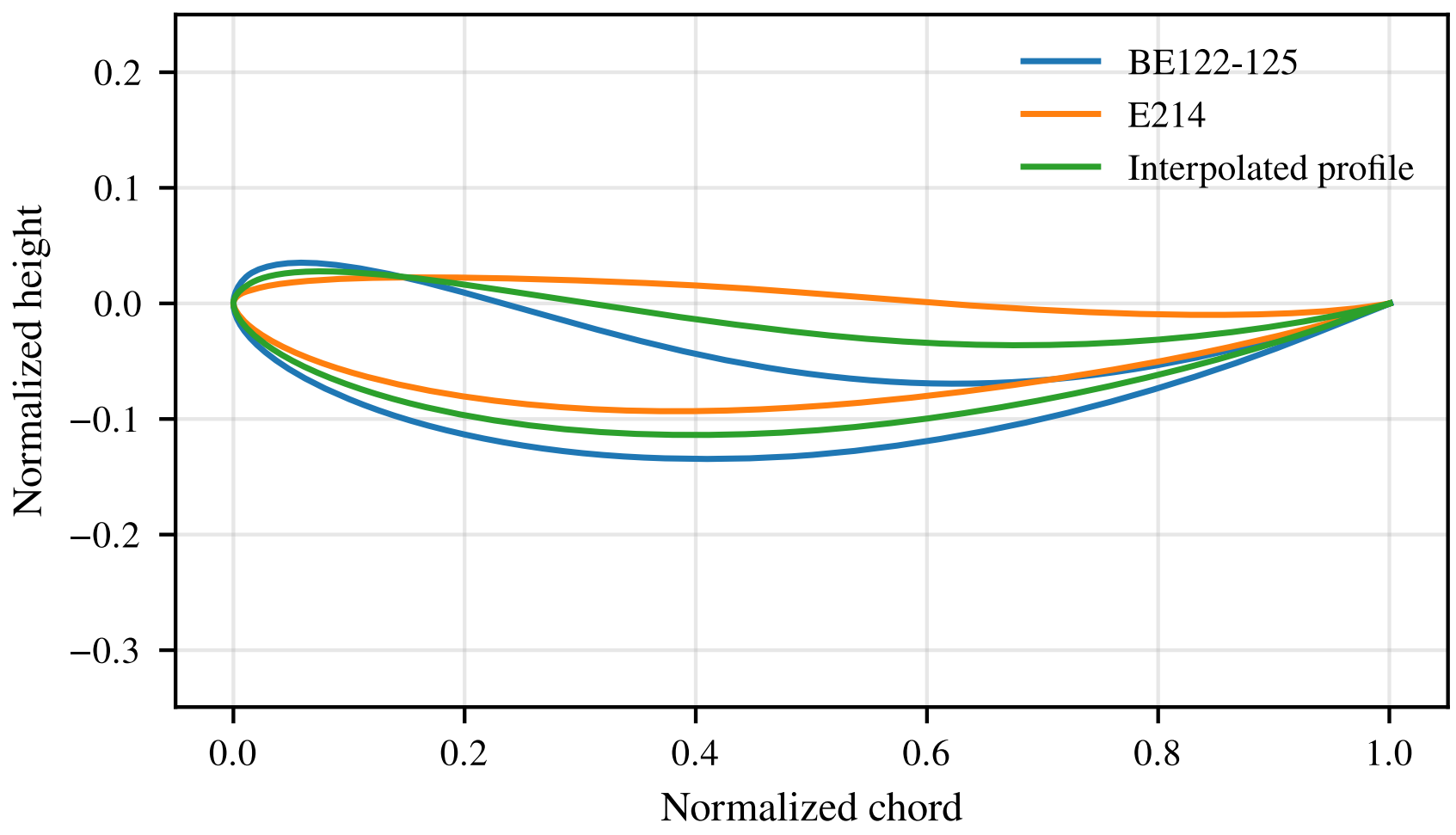}
    \caption{Interpolation between BE122-125 and E214.}
    \label{fig:interpolated_profile}
\end{figure}

Because in real applications the profile of a wing is often a customized version of a commonly available one, the profile pool was enlarged by interpolating between the named profiles using XFOIL's \cite{xfoil} airfoil interpolation feature. Pairs of profiles were combined with an interpolation weight of 0.5, producing a profile that is a simple coordinate average of the two inputs. Figure~\ref{fig:interpolated_profile} shows one such blend. This method produced the 13 blends between BE122-125 and each of the other 13 profiles, then used to generate wing geometries together with the two named profiles that are not part of the base set of profiles. The resulting profile identifiers are listed in Table~\ref{tab:profile_ids}. Profiles $P = 13$--$27$ appear exclusively in the 20 geometries of the interpolated test set, which evaluates the models trained on profiles $P = 1$--$12$ on unseen but ``familiar'' shapes. Several of the additional profiles are combined with more than one parameter set, so that the loss of prediction quality can be attributed to the profile shape separately from the remaining wing parameters.

\begin{table}[hbt!]
\caption{\label{tab:profile_ids} The 27 airfoil profile variants of the dataset. Profiles $P = 1$--$12$ are used in the base set. Profiles $P = 13$--$27$ are used only for the interpolated set. Interpolated variants are denoted $A + B$, the coordinate average of profiles $A$ and $B$.}
\centering
\begin{tabular}{clc@{\qquad\qquad}clc}
\toprule
$P$ & Profile & Type & $P$ & Profile & Type \\
\midrule
1 & BE122-125 & named & 15 & BE122-125 $+$ E214 & interpolated \\
2 & E214 & named & 16 & BE122-125 $+$ E423 & interpolated \\
3 & E423 & named & 17 & BE122-125 $+$ FX 63-137 & interpolated \\
4 & FX 63-137 & named & 18 & BE122-125 $+$ LA2573A & interpolated \\
5 & LA2573A & named & 19 & BE122-125 $+$ LNV109A & interpolated \\
6 & LNV109A & named & 20 & BE122-125 $+$ M06-13-128 & interpolated \\
7 & M06-13-128 & named & 21 & BE122-125 $+$ MSHD & interpolated \\
8 & MSHD & named & 22 & BE122-125 $+$ NACA 2412 & interpolated \\
9 & NACA 2412 & named & 23 & BE122-125 $+$ NACA 4412 & interpolated \\
10 & NACA 4412 & named & 24 & BE122-125 $+$ NACA 6412 & interpolated \\
11 & NACA 6412 & named & 25 & BE122-125 $+$ S1223RTL & interpolated \\
12 & S1223 & named & 26 & BE122-125 $+$ S1223 & interpolated \\
13 & S1223RTL & named & 27 & BE122-125 $+$ S3021 & interpolated \\
14 & S3021 & named & & & \\
\bottomrule
\end{tabular}
\end{table}

\section{Sample Generation Details}
\label{sec:appendix_sampling}

This appendix collects the details of the sampling procedure summarized in Section~\ref{sec:sampling}: the formal statement of the space-filling property of the Sobol sequence, the CFD sweeps from which the sensitivity functions were built, the feasibility constraints, and the step-by-step generation procedure.

\begin{figure}[hbt!]
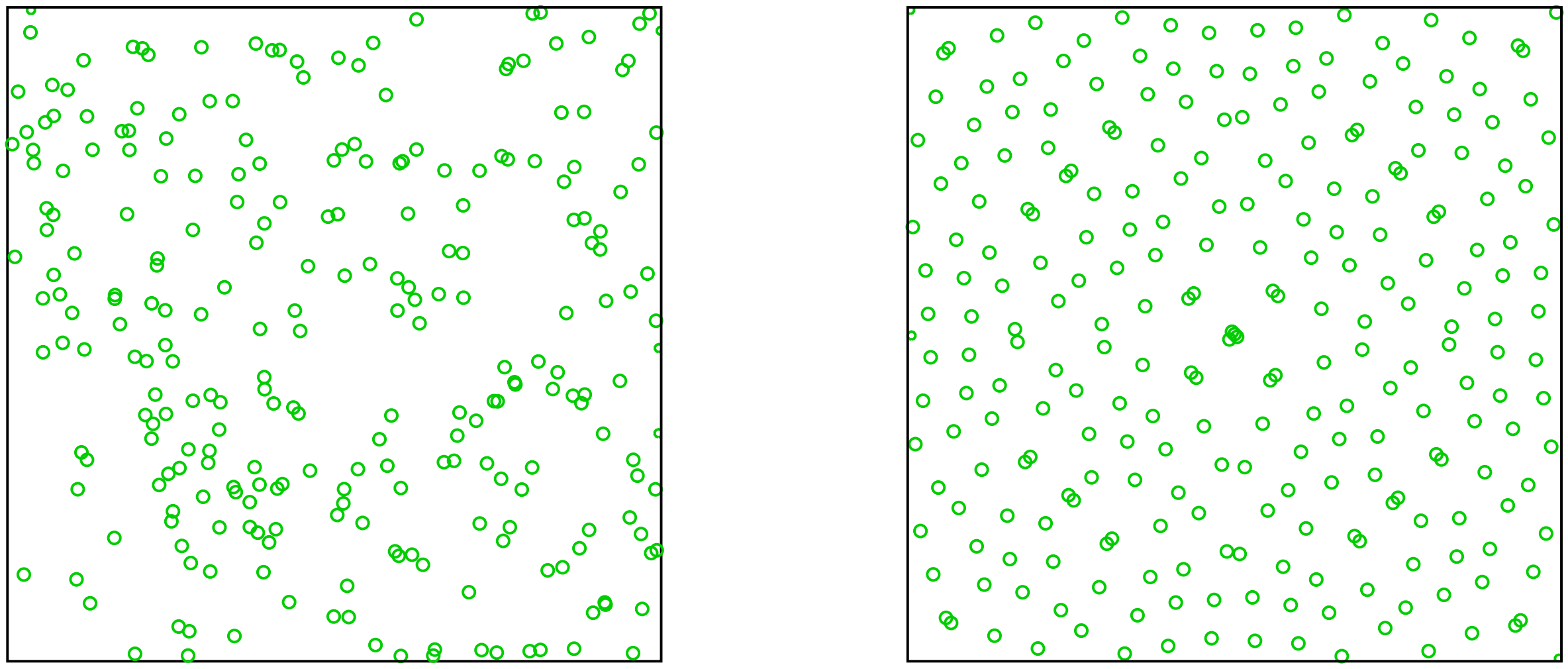

    \centering
    \subfloat[Pseudorandom sequence]{
        \includesvg[width=0.45\linewidth]{images/appendix/sampling/Pseudorandom_sequence_2D.svg}}
    \hfill
    \subfloat[Sobol sequence]{
        \includesvg[width=0.45\linewidth]{images/appendix/sampling/Sobol_sequence_2D.svg}}
    \caption{Comparison between a pseudorandom distribution and a Sobol sequence in two dimensions.}
    \label{fig:sobol}
\end{figure}

The Sobol sequence \cite{sobol} belongs to the family of quasi-random low-discrepancy sequences. Letting $I^S = [0,1]^S$ be the $S$-dimensional unit hypercube and $f$ a real integrable function over $I^S$, the objective of such a sequence is to produce points $x_i$ satisfying
\begin{equation}
    \lim_{n\rightarrow\infty} \frac{1}{n}\sum_{i=1}^{n} f(x_i) = \int_{I^S} f,
    \label{eq:sobol}
\end{equation}
with the convergence being as fast as possible. Since the equality must hold for any integrable $f$, the points have to fill $I^S$ leaving as few holes as possible, which is precisely the property required of a set of geometries meant to train a surrogate model over the whole design space. Figure~\ref{fig:sobol} compares a pseudorandom distribution with a Sobol sequence of the same size in two dimensions: the clusters and voids of the former are absent from the latter.

The sensitivity functions $S(p)$ of Eq.~\eqref{eq:sensitivity} were built from univariate CFD sweeps about a baseline geometry with $\alpha_c = \theta_t = \SI{0}{\degree}$, $c_c = \SI{600}{\milli\meter}$, $c_t = \SI{455}{\milli\meter}$, and $\Lambda = \SI{0}{\degree}$, using the solver setup of Section~\ref{sec:cfd}. The values at which each parameter was tested are listed in Table~\ref{tab:sensitivity_test_values}. They span the full range of each parameter. Cubic splines were fitted through the resulting $C_l(p)$ and $C_d(p)$ curves and differentiated analytically to obtain the two derivatives of Eq.~\eqref{eq:sensitivity}.

\begin{table}[hbt!]
\caption{\label{tab:sensitivity_test_values} Parameter values simulated to build the sensitivity functions of Eq.~\eqref{eq:sensitivity}. Each parameter was swept individually, with all others held at their baseline value.}
\centering
\begin{tabular}{llc}
\toprule
Parameter & Tested values & Unit \\
\midrule
$\alpha_c$  & $-10, -8, -6, -4, -2, 0, 2, 4, 5$          & \si{\degree} \\
$\theta_t$  & $-10, -8, -6, -4, -2, 0, 2, 4, 6, 8, 10$   & \si{\degree} \\
$c_c$       & $550, 575, 600, 625, 650, 675, 700$        & \si{\milli\meter} \\
$c_t$       & $350, 400, 455, 500, 550, 600, 650, 700$   & \si{\milli\meter} \\
$\Lambda$   & $-8.5, -4, 0, 4, 8, 12, 16, 20$            & \si{\degree} \\
\bottomrule
\end{tabular}
\end{table}

The geometric parameters do not act independently of one another: they combine into a single shape, which may be infeasible even when every individual parameter lies within its bounds. Two families of constraints, described in Section~\ref{sec:sampling}, were therefore imposed. Equation~\eqref{eq:constraints_aoa} prevents flow separation due to excessive AoA values on the wing tips, Eq.~\eqref{eq:constraints_fw_disp} bounds the forward and rearward extent of the centerline section, and Eq.~\eqref{eq:constraints_tip} additionally limits the tip chord. The symbols are those of Figs.~\ref{fig:geom_angles} and~\ref{fig:geom_box}. Every candidate violating any of Eqs.~\eqref{eq:constraints} is discarded before the weighted selection.

\begin{subequations}
    \label{eq:constraints}
    \begin{gather}
        \SI{-10}{\degree} < \alpha_c + \theta_t < \SI{5}{\degree}, \label{eq:constraints_aoa}\\
        \SI{350}{\milli\meter} \leq c_c - s\tan\Lambda \leq \SI{700}{\milli\meter}, \label{eq:constraints_fw_disp}\\
        c_t - c_c + s\tan\Lambda \leq \SI{700}{\milli\meter}, \label{eq:constraints_tip}
    \end{gather}
\end{subequations}

\subsection{Complete Sample Generation Procedure}

The complete sample generation procedure, implemented as a single Python script, is given as Algorithm~\ref{alg:sampling}.

\begin{algorithm}[hbt!]
\caption{Sensitivity- and coverage-weighted sample generation}
\label{alg:sampling}
\KwIn{parameter bounds (Table~\ref{tab:params}), sensitivity functions $S_i(p)$, rounding bases $\Delta_i$, target sample count $N$, coverage weight $w_c$}
\KwOut{$N$ selected parameter vectors with assigned profiles}
$M \gets M_0$\;
\Repeat{$|\mathcal{C}| \geq 3N$}{
  generate a scrambled 5D Sobol sequence of $M$ points and rescale each axis to the corresponding parameter range (enlarged by $\Delta_i/2$)\;
  round each coordinate to the nearest multiple of $\Delta_i$; remove duplicates\;
  $\mathcal{C} \gets$ candidates satisfying Eqs.~\eqref{eq:constraints}\;
  \If{$|\mathcal{C}| < 3N$}{$M \gets 2M$}
}
draw $p_1 \in \mathcal{C}$ with probability $\propto W_j = \prod_{i=1}^{5} S_i(p_{i,j})$\;
$\mathcal{S} \gets \{p_1\}$; $\mathcal{C} \gets \mathcal{C} \setminus \{p_1\}$\;
\For{$k \gets 2$ \KwTo $N$}{
  \ForEach{$p_j \in \mathcal{C}$}{
    $W_j \gets \prod_{i=1}^{5} S_i(p_{i,j})$\;
    $d_j \gets \min_{p \in \mathcal{S}} \lVert p_j - p \rVert$\;
    $\sigma_j \gets w_c\,d_j/\max_k d_k + (1-w_c)\,W_j/\max_k W_k$ \tcp*{Eq.~\eqref{eq:selection_score}}
  }
  $j^\ast \gets \arg\max_j \sigma_j$\;
  $\mathcal{S} \gets \mathcal{S} \cup \{p_{j^\ast}\}$; $\mathcal{C} \gets \mathcal{C} \setminus \{p_{j^\ast}\}$\;
}
assign each $p \in \mathcal{S}$ a profile via a separate scrambled 1D Sobol sequence rescaled to $[0.5, 12.5]$ and rounded to the nearest integer\;
export $\mathcal{S}$ (parameters + profile) to the CATIA-driving database\;
\end{algorithm}

The rounding bases $\Delta_i$ are $(\alpha_c, \theta_t, c_c, c_t, \Lambda) = (\SI{0.5}{\degree}, \SI{0.5}{\degree}, \SI{5}{\milli\meter}, \SI{5}{\milli\meter}, \SI{1}{\degree})$. Each range is enlarged by half its rounding base before rescaling so that the rounding does not deplete the two outermost bins relative to the interior ones. The $3N$ survivor threshold of the constraint-filtering loop keeps the following weighted selection meaningful. The sensitivity weight $W_j$ is a product over the five sensitivity functions, so a candidate is rated informative only if it is informative in every dimension simultaneously. The coverage term $d_j$ is the Euclidean distance to the nearest already-selected sample and is therefore recomputed at every draw. Both terms are normalized by their maximum over the remaining candidates before being combined into the score
\begin{equation}
    \sigma_j = w_c\,\frac{d_j}{\max_k d_k} + (1 - w_c)\,\frac{W_j}{\max_k W_k},
    \label{eq:selection_score}
\end{equation}
of which the maximum is selected at each iteration; the first sample is instead drawn at random with probability proportional to $W_j$, since no distance can be defined for an empty selection. The coverage weight was set to $w_c = 0.5$, giving the two criteria equal importance, to prevent the selection from collapsing onto the most sensitive corners of the design space. The selected samples are finally written to a \texttt{.csv} file, which is read by CATIA to drive the parametric model and produce the CAD geometries used for the CFD simulations.

The procedure yields the 174 geometries from the stratified split reported in Section~\ref{sec:sampling}, whose parameter distributions are those shown in Fig.~\ref{fig:param_dist}. The 20 geometries of the interpolated split, built on profiles $P = 13$--$27$, were generated with the same parameter sampling.

\subsection{Sensitivity Analysis on an In-Dataset Profile} The sensitivity functions used above were built from a sweep that used a proprietary internal wing profile, which is not part of the released profile pool of Appendix~\ref{sec:appendix_profiles} and cannot be shared (see the excluded-data discussion of Section~\ref{sec:reproducibility}). To confirm that the resulting sensitivity functions are still representative of the profiles actually distributed with the dataset, the same univariate sweep, with the identical baseline and ride height, was repeated on profile 12 (S1223). Unlike the spline-based derivative of Section~\ref{sec:sampling}, this one-off validation estimates $\partial C_l/\partial p$ and $\partial C_d/\partial p$ directly at the sweep's tested values with a nonuniform-grid, second-order-accurate three-point (central) finite difference.

\begin{figure}[hbt!]
    \centering
    \includesvg[width=\linewidth]{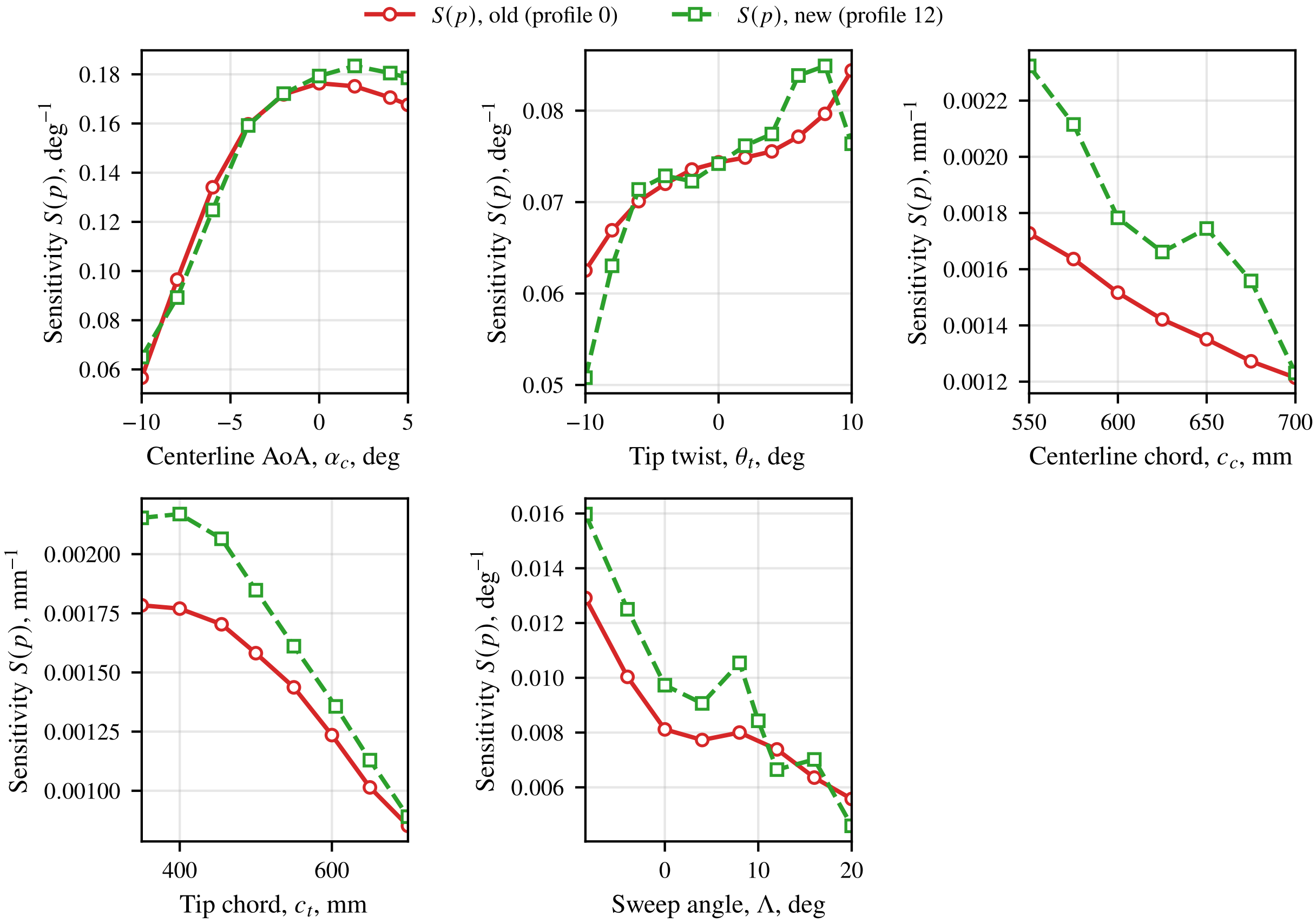}
    \caption{Sensitivity function $S(p)$ of Eq.~\eqref{eq:sensitivity} for each continuous parameter, computed from the original internal profile sweep (red) and from an equivalent repeat sweep on profile 12 (S1223, green dashed), both via the central-difference estimate described above.}
    \label{fig:new_sensitivity_comparison}
\end{figure}

Figure~\ref{fig:new_sensitivity_comparison} compares the resulting $S(p)$ for the two profiles across all five parameters. The Pearson correlation between the two curves ranges from $0.909$ ($\theta_t$) to $0.997$ ($c_t$) across the five parameters, and the relative $L_2$ difference between them ranges from $4\%$ ($\alpha_c$) to $24\%$ ($c_c$). Most importantly for the sampling procedure, the resulting importance ranking of the five parameters by mean $S(p)$ ($\alpha_c$, $\theta_t$, $\Lambda$, $c_c$, $c_t$, from most to least sensitive) is identical for both profiles, indicating that the sensitivity-weighted sampling of Section~\ref{sec:sampling} would have concentrated its samples in the same regions of the design space regardless of which profile the baseline sweep had used.

\section{Interaction of Geometric Parameters}
\label{sec:appendix_interaction}

The sensitivity-weighted sampling strategy of Section~\ref{sec:sampling} relies on sensitivity functions $S(p)$ of Eq.~\eqref{eq:sensitivity} computed by varying one parameter at a time, which implicitly assumes that the influence of each geometric parameter on the aerodynamic coefficients is independent of the others, i.e., that the total change in $C_l$ or $C_d$ can be approximated by the linear superposition of the individual marginal effects. To evaluate this assumption, a post hoc analysis was performed on the subset of the dataset at the \SI{45}{\milli\meter}/\SI{45}{\milli\meter} ride height, isolating geometric dependencies from ground-effect variations.

\begin{figure}[hbt!]
    \centering
    \includesvg[width=\linewidth]{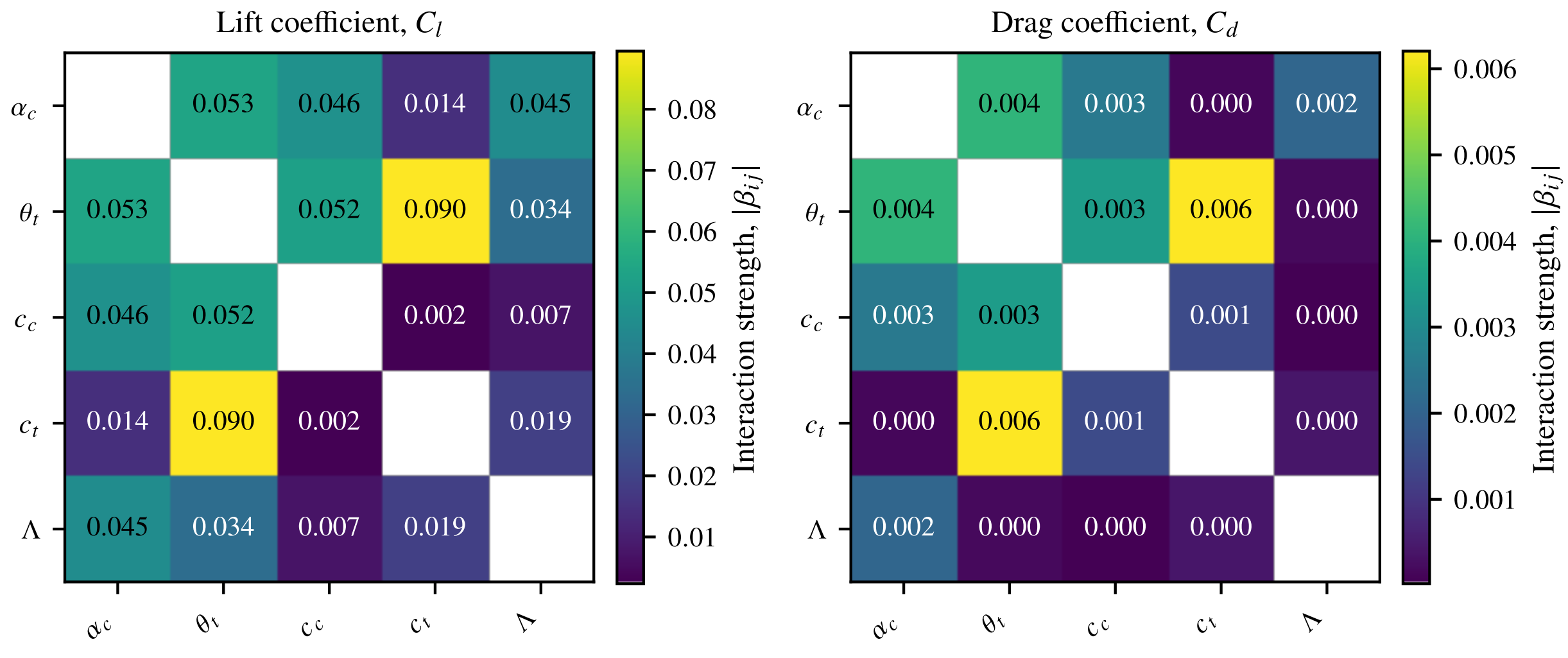}
    \caption{Interaction strength matrices $|\beta_{ij}|$ for $C_l$ (left) and $C_d$ (right) at the \SI{45}{\milli\meter} ride height.}
    \label{fig:interaction_heatmaps}
\end{figure}

A polynomial regression model with interaction terms was fitted to the standardized continuous parameters of Table~\ref{tab:params} ($\alpha_c$, $\theta_t$, $c_c$, $c_t$, $\Lambda$, normalized to zero mean and unit variance). The profile parameter $P$ is excluded from the analysis: as discussed in Section~\ref{sec:sampling}, it is a categorical index rather than a continuously variable dimension, so no sensitivity function is defined for it and a product term between its standardized index and another parameter has no physical meaning. Letting $\mathbf{x}\in\mathbb{R}^n$ denote this parameter vector and $y$ the target coefficient ($C_l$ or $C_d$), the response surface is
\begin{equation}
    \hat{y} = \beta_0 + \sum_{i=1}^{n} \beta_i x_i + \sum_{i=1}^{n}\sum_{j>i}^{n} \beta_{ij} x_i x_j,
    \label{eq:interaction_poly}
\end{equation}
where $\beta_i$ is the linear effect of parameter $i$ and the interaction strength $I_{ij} = |\beta_{ij}|$ quantifies how strongly the effect of parameter $i$ depends on parameter $j$. A high $I_{ij}$ indicates a violation of the independence assumption used during sampling. Figure~\ref{fig:interaction_heatmaps} shows the results of this analysis.

The two coefficients are dominated by the same pair but differ in overall magnitude. For $C_l$, the strongest coupling is between the tip twist $\theta_t$ and the tip chord $c_t$ ($I \approx 0.090$). This is consistent with the local lift increment produced by twist scaling with the chord over which it acts. It suggests the independence assumption led to some undersampling where a large tip chord combines with extreme tip twist. For $C_d$, the same $\theta_t$--$c_t$ pair remains strongest, but interaction strengths are roughly an order of magnitude smaller overall (maximum $I \approx 0.006$), indicating that individual parameter contributions to drag are largely additive and that the univariate sensitivity functions used for sampling remain an accurate basis for this coefficient.

These results suggest that future iterations of the sampling strategy could augment $S(p)$ with a term proportional to the cross-parameter interaction strengths $I_{ij}$, biasing sample selection toward regions where highly interacting parameter pairs (e.g., $\theta_t$ and $c_t$) are simultaneously near their bounds, in order to better capture the nonlinear aerodynamic phenomena in those configurations.

\section{Ride-Height Sweep and Choking Assessment}
\label{sec:appendix_ride_height}

The efficiency of a downforce-generating wing increases as its distance from the ground decreases: the gap between the wing and the ground acts as a Venturi channel, accelerating the flow and further lowering the pressure on the suction side, so that downforce grows considerably faster than drag. This behavior has a limit \cite{inverted_wing}. Below a critical clearance, the pressure beneath the wing becomes so low that the flow can no longer recover to ambient conditions before the trailing edge and it separates. This causes the mass flow between the wing and the ground to become choked, and the wing suffers a sharp increase in drag together with a stark loss of downforce. Correia et al.~\cite{inverted_wing} test this effect on a single-element inverted wing, showing that its downforce peaks at a height-to-chord ratio of $h/c \approx 0.09$ and falls off below it. This dataset's lowest clearance corresponds to $h/c_c = 0.145$ for a centerline chord of \SI{600}{\milli\meter}. Therefore, it had to be established that none of the simulated configurations fall into this nonlinear regime, which the RANS setup of Section~\ref{sec:cfd} would not resolve reliably, and which a surrogate model trained predominantly on well-behaved configurations could not be expected to predict.

The verification was carried out during the development of the CFD methodology, before the parametric model was frozen, and therefore uses a fixed development geometry rather than one of the sampled geometries of the dataset. That geometry was simulated at 11 flat ride heights between \SI{30}{\milli\meter} and \SI{100}{\milli\meter}, thus covering the four ride heights of Table~\ref{tab:params}. Each configuration was obtained by translating the wing rigidly in the vertical direction and regenerating the mesh, and was solved with the same solver, turbulence model, and freestream conditions as the dataset simulations. Only flat setups are considered, for the reasons given in Appendix~\ref{sec:appendix_geometry}.

\begin{figure}[hbt!]
    \centering
    \includesvg[width=\linewidth]{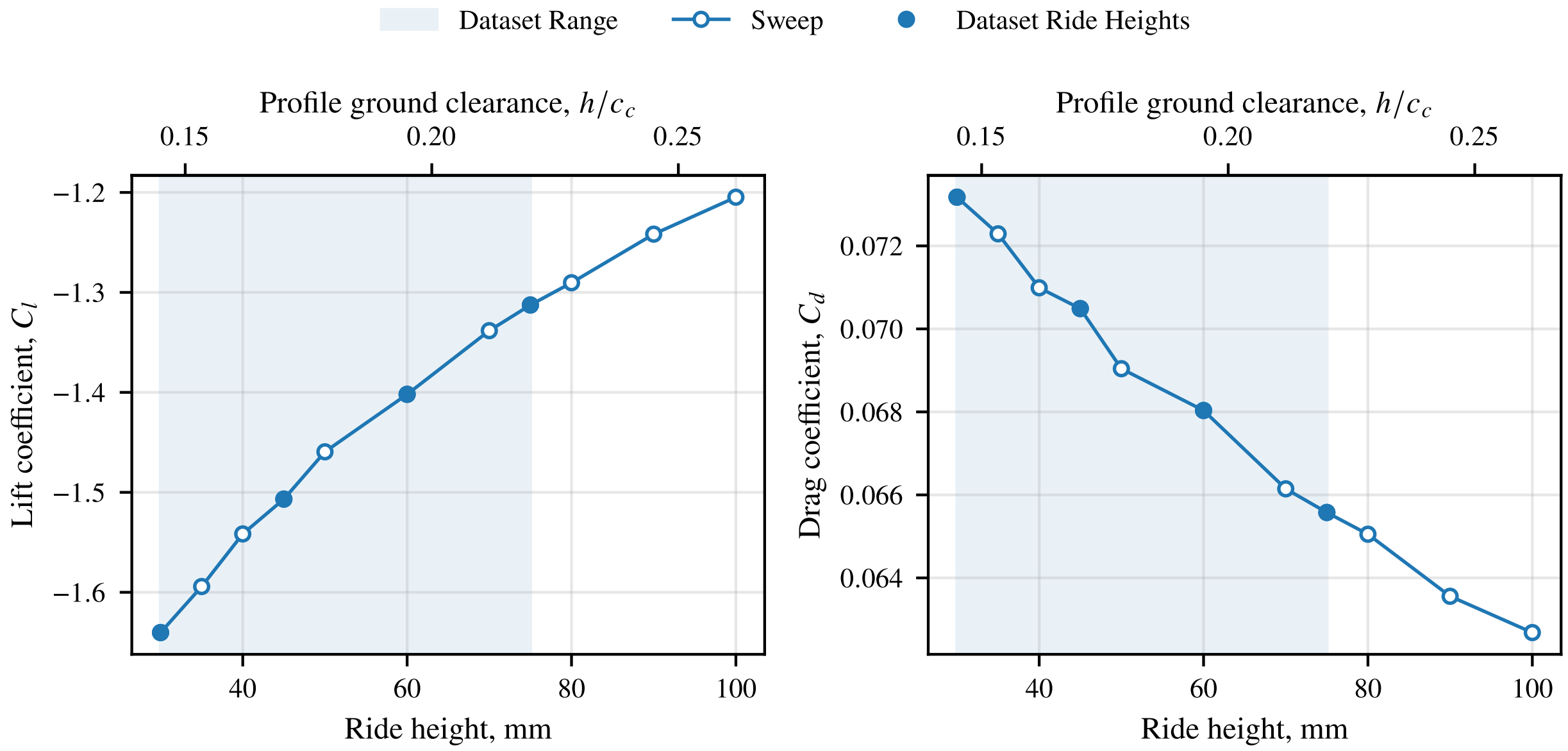}
    \caption{Aerodynamic coefficients of the development geometry over the swept ride heights. The shaded band marks the interval covered by the dataset, whose four ride heights are highlighted.}
    \label{fig:ride_height_sweep}
\end{figure}

Figure~\ref{fig:ride_height_sweep} reports the resulting coefficient trends, against both the ride height, measured at the lower edge of the endplate, and the corresponding normalized ground clearance $h/c_c$, defined with $h$ as the profile clearance. Both coefficients vary smoothly and monotonically over the entire range. This is the expected unchoked ground-effect behavior, and it confirms that the four ride heights of the dataset, and the whole interval between them, lie in the linear operating region. The absolute coefficient values of Fig.~\ref{fig:ride_height_sweep} belong to the development geometry and are not directly comparable to those of the dataset samples.

\section{Dataset Splitting Procedure}
\label{sec:appendix_splits}

This appendix states the splitting procedure summarized in Section~\ref{subsec:splits}. Two strategies were used: iterative stratification, which produced the stratified split, and drag-outlier allocation, which produced the drag-outlier split used to probe the edges of the aerodynamic envelope. Both operate on the 174 unique base-set geometries rather than on the 690 individual simulations, and both reattach every simulated ride height of a geometry to whichever subset that geometry was assigned to. The target proportions are 80\% training, 10\% validation, and 10\% test in both cases.

The stratifying problem is recast as a multilabel one and solved with the iterative stratification algorithm of Ref.~\cite{iterstrat}, which allocates samples one label at a time, always serving the label that is currently scarcest in the subset that is furthest from its quota. The procedure is as follows:

\begin{enumerate}
    \item \textbf{Label construction}: The five continuous parameters of Table~\ref{tab:params} are each discretized into 15 uniformly spaced bins, and the resulting bin indices are one-hot encoded together with the categorical profile parameter $P$. Every geometry is thereby turned into a binary indicator vector over the union of all parameter bins and all profiles.
    \item \textbf{First split}: Iterative stratification is applied to the full set of geometries with a test fraction of $0.2$, yielding the training set and a held-out remainder of 20\%.
    \item \textbf{Second split}: The same algorithm is applied to the remainder with a test fraction of $0.5$, dividing it into the validation and test sets.
    \item \textbf{Ride-height reattachment}: Each geometry contributes its run identifiers, one per simulated ride height (Table~\ref{tab:params}), to the subset it was assigned to, and the three lists of run identifiers are written to a configuration file that the training pipelines read directly.
\end{enumerate}

The drag-outlier split replaces steps 1--3 while keeping step 4 unchanged. Its purpose is to place the extremes of the drag distribution deliberately in the validation and test sets. Given $n$ geometries and validation and test fractions $v = t = 0.1$:
\begin{enumerate}
    \item \textbf{$\bm{C_d}$ Ranking}: The geometries are ranked by their drag coefficient at the \SI{45}{\milli\meter} ride height, and the $n \times (v+t) \div 2$ most extreme of them, rounded down to an even number and taken half from each end of the ranking, are marked as outliers.
    \item \textbf{Outlier Selection}: The outliers are re-sorted by their absolute deviation from the median drag coefficient, most extreme first, and dealt alternately to the test and validation sets, with the test set receiving the most extreme geometry.
    \item \textbf{Random Fill}: The remaining geometries are randomly shuffled and used to fill the rest of the validation and test quotas, while whatever is left forms the training set.
\end{enumerate}

\section{Model Configurations}
\label{sec:appendix_models}

This appendix documents the configuration each of the six models was trained with. Tables~\ref{tab:abupt_config}, \ref{tab:geotransolver_config}, \ref{tab:aerojepa_config}, \ref{tab:gaot_config}, \ref{tab:domino_config}, and \ref{tab:xmeshgraphnet_config} cover the architecture-specific choices that have no counterpart in the other models. As stated in Section~\ref{subsec:models}, the architecture-specific settings were determined by a combination of Optuna~\cite{optuna} search and empirical testing on the validation set of the stratified split, whereas the number of epochs and the learning-rate settings were kept at the values found in each model's reference implementation. The released dataset contains the \texttt{.yaml} configuration files used for each model.

\begin{table}[H]
\caption{\label{tab:abupt_config} AB-UPT configuration.}
\centering
\begin{tabularx}{\textwidth}{lX}
\toprule
\multicolumn{2}{l}{\textit{Architecture}} \\
\midrule
Hidden dimension & 192 (all blocks) \\
Attention heads & 3 \\
Multilayer perceptron (MLP) hidden dimension & 768 \\
Position embeddings & Rotary \\
Geometry branch depth & 1 block \\
Supernode pooling & \num{20000} $\to$ \num{2000} points (radius $0.25$, normalized coordinates) \\
Physics trunk & 6 blocks (1 perceiver + 5 self-attention) \\
Surface-domain decoder & 6 blocks \\
Cross-attention / volume decoder & None (volume branch disabled) \\
Attention anchors & \num{2000} surface points \\
\bottomrule
\end{tabularx}
\end{table}

\begin{table}[H]
\caption{\label{tab:geotransolver_config} GeoTransolver configuration.}
\centering
\begin{tabularx}{\textwidth}{lX}
\toprule
\multicolumn{2}{l}{\textit{Architecture}} \\
\midrule
Transformer blocks & 13 \\
Width & 256 \\
Attention heads & 4 \\
MLP ratio & 2 \\
Dropout & 0.05 \\
Physics slices per block & 128 \\
Ball-query radii & [$0.045$, $0.25$, $1.25$, $6.5$] \\
Ball-query neighbors & [4, 10, 25, 64] \\
Local feature channel width & 64 \\
Sampled points & \num{100000} surface, \num{50000} geometry \\
\bottomrule
\end{tabularx}
\end{table}

\begin{table}[H]
\caption{\label{tab:aerojepa_config} AeroJEPA configuration.}
\centering
\begin{tabularx}{\textwidth}{lX}
\toprule
\multicolumn{2}{l}{\textit{Architecture}} \\
\midrule
Token dimension & 128 \\
Point tokens & Up to 512 (farthest-point-sampling tokenizer, cluster size 16) \\
Context / target encoder, decoder & 6 transformer layers, 4 heads (each) \\
JEPA predictor head & 6 layers \\
Sampled points & \num{8192} surface (geometry input, target encoder, query set) \\
\bottomrule
\end{tabularx}
\end{table}

\begin{table}[H]
\caption{\label{tab:gaot_config} GAOT configuration.}
\centering
\begin{tabularx}{\textwidth}{lX}
\toprule
\multicolumn{2}{l}{\textit{Architecture}} \\
\midrule
Latent token grid & $64\times32\times32$ \\
Graph neural operator (GNO) radius & 0.033 \\
Attention & Cosine \\
Projection width & 256 \\
Node sampling ratio & 0.7 \\
Geometry embedding & Statistical local shape descriptor (encoder side only) \\
ViT processor & 10 layers, hidden size 256, 8 heads\\
\bottomrule
\end{tabularx}
\end{table}

\begin{table}[H]
\caption{\label{tab:domino_config} DoMINO configuration.}
\centering
\begin{tabularx}{\textwidth}{lX}
\toprule
\multicolumn{2}{l}{\textit{Architecture}} \\
\midrule
Mode & Surface-only \\
Geometry grid & $64\times32\times32$, processed by a UNet (8 base filters) \\
Multiscale point-conv radii & [$0.01$, $0.05$, $0.1$, $0.5$, $1.0$] \\
Multiscale point-conv neighbors & [8, 16, 24, 30, 36] \\
Basis-function / aggregation networks & Width 256, Fourier features (20 / 10 modes) \\
Sampled points & \num{300000} geometry, \num{16384} supervised surface (7 neighbors/point) \\
\bottomrule
\end{tabularx}
\end{table}

\begin{table}[H]
\caption{\label{tab:xmeshgraphnet_config} X-MeshGraphNet configuration.}
\centering
\begin{tabularx}{\textwidth}{lX}
\toprule
\multicolumn{2}{l}{\textit{Architecture}} \\
\midrule
Graph resolutions & \num{10000}, \num{20000}, \num{40000} nodes \\
Node degree & 6 \\
Partitions per geometry & 2 \\
Node placement & Curvature-driven (curvature weight 0.3) \\
Edge filtering & \SI{120}{\degree} normal threshold, \SI{45}{\degree} edge-angle threshold \\
Message-passing steps & 15 \\
Hidden dimension & 256 \\
Activation & Gaussian error linear unit (GELU) \\
\bottomrule
\end{tabularx}
\end{table}

\section{AB-UPT Trained with Volumetric Targets}
\label{sec:appendix_abupt_volume}

AB-UPT is the only architecture in the benchmark whose surface and volume branches are coupled by default with cross-attention, and a single-domain model is not directly available in the reference code. It was therefore also trained against both the surface fields and the volumetric $C_p$ and $\bm{u}$ fields. The benchmark of Section~\ref{sec:benchmark} does not use this capability. Every model there is trained on the surface fields alone, so that no architecture is supervised on target fields the others never see. This appendix reports the volume-trained variant separately, as additional information on what the extra supervision costs or buys on the surface task.

The two configurations differ only along the volume path. The volume-trained variant instantiates a volume-domain decoder of six blocks alongside the surface one, replaces two of the five self-attention blocks of the physics trunk with cross-attention blocks so that the surface and volume branches exchange information, and evaluates attention against \num{10000} volume anchor points in addition to the \num{2000} surface anchors. Every other setting is shared with the surface-only run of Appendix~\ref{sec:appendix_models}. Both variants were trained on the stratified and on the drag-outlier split and evaluated on all three test sets under the identical protocol, so Table~\ref{tab:abupt_targets} compares training targets and nothing else.

\begin{table}[hbt!]
\caption{\label{tab:abupt_targets} AB-UPT trained on the surface fields alone against the variant additionally trained on the volumetric targets, on all three test sets. Both variants are evaluated on the surface fields only.}
\centering
\begin{tabular}{llcccc}
\toprule
Test set & Training targets & $C_l$ Err. ($\mu, \sigma$), \% & $C_d$ Err. ($\mu, \sigma$), \% & $C_p$ $L_2$, \% & $\lvert\bm{w_s}\rvert$ $L_2$, \% \\
\midrule
Stratified & Surface only & 0.85, 0.89 & 0.89, 0.74 & 2.63 & 2.46 \\
 & Surface + volume & 0.62, 0.64 & 1.04, 0.86 & 2.52 & 2.36 \\
\midrule
Drag-outlier & Surface only & 2.82, 4.15 & 1.92, 1.96 & 4.18 & 3.76 \\
 & Surface + volume & 4.54, 6.55 & 3.71, 2.72 & 5.26 & 4.39 \\
\midrule
Interpolated & Surface only & 1.56, 1.25 & 4.51, 3.58 & 5.00 & 3.85 \\
 & Surface + volume & 1.86, 1.46 & 4.90, 3.93 & 5.33 & 3.78 \\
\bottomrule
\end{tabular}
\end{table}

The additional supervision costs accuracy on all dataset splits. This is a change one would expect: the volumetric term consumes part of a fixed model capacity and a fixed optimization budget on a task the surface metrics do not reward.

The surface accuracy reported for AB-UPT in Table~\ref{tab:results} is therefore a property of the architecture rather than a byproduct of extra supervision that the other five models never receive. Consequently, restricting every model to the surface fields costs the comparison nothing and removes an asymmetry that would otherwise be impossible to interpret.

The volumetric fields themselves are not evaluated in this paper. The variant is characterized here purely by what it was trained on and by its surface accuracy. Quantifying volumetric prediction error on this dataset is left to future work.

\newpage
\pagebreak

\section*{Acknowledgments}
The authors acknowledge BMW M Motorsport and BMW Group IT for providing computational resources and technical support. They acknowledge NVIDIA Corporation for technical support with their PhysicsNeMo \cite{physicsnemo} software suite, and Siemens AG for providing access to \href{https://www.siemens.com/en-us/products/simcenter/fluids-thermal-simulation/star-ccm/}{STAR-CCM+} licenses.
A.~Cenzato acknowledges support from RWTH Aachen University. AI assistance (an LLM-based coding agent) was used to help draft \LaTeX{} formatting, generate the data-analysis notebooks and figures, and adapt prose from the authors' prior internal reports into this manuscript. All technical content, figures, and claims were reviewed and verified by the authors against the underlying dataset and simulation results.

\bibliography{main}

\end{document}